\documentclass[nohyperref]{article}

\usepackage{microtype}
\usepackage{graphicx}
\usepackage{subfigure}
\usepackage{booktabs} % for professional tables
\usepackage{rotating}
\usepackage[table,xcdraw]{xcolor}
\usepackage{multirow}
\usepackage{svg}
\usepackage{tablefootnote}
\usepackage{threeparttable}
\usepackage{hyperref}
\usepackage{graphicx}
\usepackage{float}
\usepackage{soul}
\hypersetup{colorlinks=true, allcolors=blue}

\usepackage{hyperref}

\usepackage[accepted]{icml2022}

\usepackage{amsmath}
\usepackage{amssymb}
\usepackage{mathtools}
\usepackage{amsthm}

\usepackage[capitalize,noabbrev]{cleveref}

\theoremstyle{plain}

\theoremstyle{definition}

\theoremstyle{remark}

\usepackage[textsize=tiny]{todonotes}

\icmltitlerunning{manuscript of paper accepted to CFOL Workshop @ ICML 2022}

\newif\ifanonymize
\anonymizefalse

\ifanonymize
\newcommand*{\metalearningwebsite}{{\textit{Anonymous-website}}}
\newcommand*{\wcci}{{\textit{Anonymous-conference}}}
\newcommand*{\automl}{\textit{Anonymous-conference2}}
\newcommand*{\outputlink}{{\textit{Anonymous-URL}}}
\newcommand*{\firstcodalabpage}{{\textit{Anonymous-competition-page}}}
\newcommand*{\automlchallenge}{{\textit{Anonymous-challenge}}}
\newcommand*{\metaReveal}{{(\textit{Anonymous-paper})}}
\newcommand*{\MoRiHa}{{\textit{Anonymous-team-01}}}
\newcommand*{\neptune}{{\textit{Anonymous-team-02}}}
\newcommand*{\AIpert}{{Anonymous-team-03}}
\newcommand*{\automlfreiburg}{{\textit{Anonymous-team-04}}}
\newcommand*{\automlhannover}{{\textit{Anonymous-team-05}}}
\newcommand*{\amsks}{{\textit{Anonymous-team-05}}}
\newcommand*{\pprp}{{\textit{Anonymous-team-06}}}
\newcommand*{\arushsharma}{{\textit{Anonymous-team-07}}}
\newcommand*{\Xavier}{{\textit{Anonymous-team-08}}}

\newcommand*{\firstteam}{{\textit{Anonymous-URL}}}
\newcommand*{\secondteam}{{\textit{Anonymous-URL}}}
\newcommand*{\thirdteam}{{\textit{Anonymous-URL}}}
\newcommand*{\fourthteam}{{\textit{Anonymous-URL}}}
\newcommand*{\fifthteam}{{\textit{Anonymous-URL}}}

\newcommand*{\codalab}{{\textit{Anonymous-platform}}}

\else

\newcommand*{\metalearningwebsite}{\url{https://metalearning.chalearn.org/}}
\newcommand*{\wcci}{{WCCI 2022}}
\newcommand*{\automl}{{AutoML-Conf 2022}}
\newcommand*{\outputlink}{{\href{https://drive.google.com/drive/folders/1HrMZNQeR1kDOmLqjRGgcozut2M_k-jHs?usp=sharing}{[link]}}}
\newcommand*{\firstcodalabpage}{{\url{https://codalab.lisn.upsaclay.fr/competitions/753}}}
\newcommand*{\automlchallenge}{{AutoML challenge \cite{Guyon2019}}}
\newcommand*{\metaReveal}{{\cite{nguyen2021}}}
\newcommand*{\MoRiHa}{{MoRiHa}}
\newcommand*{\neptune}{{neptune}}
\newcommand*{\AIpert}{{AIpert}}
\newcommand*{\automlfreiburg}{{automl-freiburg}}
\newcommand*{\automlhannover}{{automl-hannover}}
\newcommand*{\amsks}{{amsks}}
\newcommand*{\pprp}{{pprp}}
\newcommand*{\arushsharma}{{arushsharma24}}
\newcommand*{\Xavier}{{Xavier}}
\newcommand*{\firstteam}{{\href{https://github.com/fmohr/meta-learn-from-LC-2022}{[CODE URL]} \href{https://drive.google.com/file/d/1CdLOEuM-7C94WdTSwHHgelGAOfZ-xCNW/view?usp=sharing}{[FACTSHEET]}}}
\newcommand*{\secondteam}{{\href{https://github.com/neptuneai/MetaDL}{[CODE URL]} \href{https://drive.google.com/file/d/1eAI0GD-0cYwI3XND23pkX5t0nvWVRy6t/view?usp=sharing}{[FACTSHEET]}}}
\newcommand*{\thirdteam}{{\href{https://github.com/EleGo9/Meta-Learning-Curve_AIpert}{[CODE URL]} \href{https://drive.google.com/file/d/1gMSTlcD2__EkFKW3qd7bV9WYb51PWs1O/view?usp=sharing}{[FACTSHEET]}}}
\newcommand*{\fourthteam}{{\href{https://github.com/sebastianpinedaar/meta-learn-from-LC}{[CODE URL]} \href{https://drive.google.com/file/d/1y8mwxa5jVt-BM5iLPXJHVHwxpbW0OIt-/view?usp=sharing}{[FACTSHEET]}}}
\newcommand*{\fifthteam}{{\href{https://github.com/timruhkopf/Meta_challenge}{[CODE URL]} \href{https://drive.google.com/file/d/1bABinDdTtQ524QtHMyte0oYL336JvCgk/view?usp=sharing}{[FACTSHEET]}}}

\newcommand*{\codalab}{{\textit{Codalab}}}

\fi  

\begin{document}
\fancypagestyle{firstpage}
{
    \fancyhead[L]{manuscript of paper accepted to CFOL Workshop @ ICML 2022}
}

\twocolumn[
\icmltitle{Meta-learning from Learning Curves Challenge:\\ Lessons learned from the First Round and Design of the Second Round}

%  \\ (\normalsize{{manuscript of paper accepted to ``Complex Feedback in Online Learning" workshop @ ICML 2022})
 


\icmlsetsymbol{equal}{*}

\begin{icmlauthorlist}
\icmlauthor{Manh Hung Nguyen}{chalearn}
\icmlauthor{Lisheng Sun}{chalearn}
\icmlauthor{Nathan Grinsztajn}{inria,cnrs,udl}
\icmlauthor{Isabelle Guyon}{chalearn,inria,cnrs,ups}
\end{icmlauthorlist}

\icmlaffiliation{chalearn}{Chalearn, USA}

\icmlaffiliation{inria}{INRIA, France}

\icmlaffiliation{cnrs}{CNRS, France}

\icmlaffiliation{ups}{Université Paris-Saclay, France}

\icmlaffiliation{udl}{Université de Lille, France}

\icmlcorrespondingauthor{Manh Hung Nguyen}{hungnm.vnu@gmail.com}

% \begin{icmlauthorlist}
% \icmlauthor{Manh Hung Nguyen}{chalearn}
% \icmlauthor{Lisheng Sun}{chalearn}
% \icmlauthor{Nathan Grinsztajn}{inria, cnrs, udl}
% \icmlauthor{Isabelle Guyon}{cnrs, inria, ups, chalearn}
% \end{icmlauthorlist}

% \icmlaffiliation{chalearn}{Chalearn, USA}

% \icmlaffiliation{inria}{INRIA, France}

% \icmlaffiliation{cnrs}{CNRS, France}

% \icmlaffiliation{ups}{Université Paris-Saclay, France}

% \icmlaffiliation{udl}{Université de Lille, France}

\icmlkeywords{Machine Learning, ICML}

\vskip 0.3in
]




\begin{abstract}
Meta-learning from learning curves is an important yet often neglected research area in the Machine Learning community. We introduce a series of Reinforcement Learning-based meta-learning challenges, in which an agent searches for the best suited algorithm for a given dataset, based on feedback of learning curves from the environment. The first round  attracted participants both from  academia and industry. This paper analyzes the results of the first round (accepted to the competition program of \wcci), to draw insights into what makes a meta-learner successful at learning from learning curves. With the lessons learned from the first round and the feedback from the participants, we have designed the second round of our challenge with a new protocol and a new meta-dataset. The second round of our challenge is accepted at the \automl\ and currently on-going.
\end{abstract}

\section{Background and motivation}
\label{background}
\textbf{Meta-learning} has been playing an increasingly important role in Automated Machine Learning.
While it is a natural capability of living organisms, who constantly transfer acquired knowledge across tasks to quickly adapt to changing environments, artificial learning systems are still in their meta-learning “infancy”. They are only capable, so far, to transfer knowledge between very similar tasks. At a time when society is pointing fingers at AI for being wasteful with computational resources, there is an urgent need for learning systems, which \textbf{recycle their knowledge}. To achieve that goal, some research areas have been widely studied, including few-shot learning \cite{Wang2020}, transfer learning \cite{Zhuang2020}, representation learning \cite{Bengio2013}, continual learning \cite{Delange2021}, life-long learning \cite{Chen2018}, and meta-learning \cite{Vanschoren2018}. %Our aim is to progressively build complexity to challenge the Machine Learning community.
However, \textbf{meta-learning from learning curves}, an essential sub-problem in meta-learning \cite{mohr2022learning}, is still under-studied. This motivates the design of this new challenge series we are proposing.

Learning curves evaluate algorithm incremental performance improvement, as a function of training time, number of iterations, and/or number of examples. Our challenge design builds on top of previous challenges and work, which considered other aspects of the problem: meta-learning as a recommendation problem, but not from learning curves \cite{Guyon2019, liu_winning_2020} and few-shot learning \cite{metadl2021}. Analysis of past challenges revealed that top-ranking methods often involve switching between algorithms during training, including “freeze-thaw” Bayesian techniques \cite{Swersky2014}. However, previous challenge protocols did not allow evaluating such methods separately, due to inter-dependencies between various heuristics. Furthermore, we want to study the potential benefit of \textbf{learned policies}, as opposed to applying hand-crafted black-box optimization methods.

Our challenge took inspiration from MetaREVEAL \metaReveal, ActivMetal \cite{SunHosoya2018}, REVEAL \cite{sunhosoya2019}, and Freeze-Thaw Bayesian Optimization \cite{Swersky2014}. A meta-learner needs to learn to solve two problems at the same time: \textbf{algorithm selection} and \textbf{budget allocation}. We are interested in meta-learning strategies that leverage information on \textbf{partially trained algorithms}, hence reducing the cost of training them to convergence. We offer pre-computed learning curves as a function of time, to facilitate benchmarking. Meta-learners must “pay” a cost emulating computational time for revealing their next values. Hence, meta-learners are expected to learn the \textbf{exploration-exploitation trade-offs} between continuing “training” an already tried good candidate algorithm and checking new candidate algorithms.

In this paper, we first recap design of the first round of our challenge. Then, we present the first round results and compare the participants' solutions with our baselines. After the end of the first round, we have identified some limitations in our design and proposed a new design and a new meta-dataset for the second round.
 
\section {Design of the first round}
In this section, we describe the design of the first round of our challenge, including: data, challenge protocol, and evaluation. More details can be found on our competition page \footnote{\firstcodalabpage}.

\subsection{Learning curve data}
Despite the fact that learning curves are widely used, there were not many learning curve meta-datasets available in the Machine Learning community at the time of organizing this challenge. Taking advantage of 30 cross-domain datasets used in the \automlchallenge, we computed {\em de novo} learning curves (both on the validation sets and the test sets) for 20 algorithms, by submitting them to the \automlchallenge\ as post-challenge submissions. These algorithms are created from two base algorithms, Random Forest and Gradient Boosting, but with different values of the \textit{max\_features} hyperparameter. 
% \lisheng{we only changed 'max\_features', right?} 
We also provided meta-features of each dataset, such as type of learning task, evaluation metric, time budget, etc. In this first round, we considered a \textbf{learning curve as a function of time}. Each point on the learning curve corresponds to the algorithm performance at a certain time. 

In addition to the aforementioned real-world meta-dataset, we synthetically generated 4000 learning curves for the participants to practice and get familiar with our starting kit. The points on each synthetic learning curve are sampled from a parameterized sigmoid function with its hyperparameters generated from matrix factorizations. This allows us to create some hidden relationships between algorithms and datasets (i.e. some algorithms perform well particularly for some datasets). Details of how this synthetic meta-dataset was created can be found in \metaReveal. 

\subsection{Challenge protocol} 

We organized a novel two-phase competition protocol:  
\begin{itemize}
    \setlength{\itemsep}{0pt}%
    \setlength{\parskip}{0pt}
    \item \textbf{Development phase}: participants make as many submissions as they want \footnote{up to a comfortable limit that was never reached in our competition. The limit also aims to avoid overfitting the validation sets, though it has been shown that participants are usually aware not to overfit them \cite{overfitting-kaggle}}, 
    % \lisheng{"showing that the participants are aware not to overfit the validation set?" I didn't cite the paper, I think it needs more explanation. This footnote seems enough to me.}
    which are evaluated on the \textit{validation learning curves}.
    % \lisheng{If participants make a large number of submissions, is it possible to overfit the validation (learning curves) set?}
    % \hung{yes, they can, but that's why we keep the test learning curves hidden for evaluation in the final phase.}
    \item \textbf{Final test phase}: no new submissions are accepted in this phase. The last submission of each participant in the Development phase is transferred automatically to this phase. It is evaluated on the \textit{test learning curves}.
\end{itemize}

Note that the meta-datasets are never exposed to the participants in neither phase, because this is a challenge with code submission (only the participants' agent sees the data). 

This setting is novel because it is uncommon in reinforcement learning to have a separate phase for agent ``development'' and agent ``testing''. Validation learning curves are used during the development phase and test learning curves during the final phase, to prevent agent from overfitting. Moreover, we implemented the k-fold meta-cross-validation (with k=6) to reduce variance in the evaluations of the agents (i.e. 25 datasets for meta-training and 5 datasets for meta-testing). The final results are averaged over datasets in the test folds. 

During meta-training, learning curves and meta-data collected on 25 datasets are passed to the agent for meta-learning in any possible ways implemented by the agent. Then during meta-testing, one dataset is presented to the agent at a time. The agent interacts back and forth with an ``environment'', similarly to a Reinforcement Learning setting. It keeps suggesting to reveal algorithms' validation learning curves and choosing the current best performing algorithm based on observations of the partially revealed learning curves. 

% \textbf{If allowed, CITE WCCI paper [for details on the design]}

\subsection{Evaluation}
In this challenge series, we want to search for agents with high ``any-time learning" capacities, which means the ability to have good performances if they were to be stopped at any point in time. Hence, the agent is evaluated by the Area under the agents’ Learning Curve (ALC) which is constructed using the learning curves of the best algorithms chosen at each time step (validation learning curves in the Development phase, and the test learning curves in the Final phase). The computation of the ALC is explained in \metaReveal. The results will be averaged over all meta-test datasets and shown on the leaderboards. The final ranking is made according to the average test ALC.

As indicated in the competition rules, participants should make efforts to guarantee the reproducibility of their methods (e.g. by fixing all random seeds involved). In the Final Phase, all submissions were run \textbf{three times} with the same seed, and the run with the \textbf{worst performance} is used for the final ranking\footnote{The output of each run can be found in this Google Drive folder: \outputlink}. This penalizes any team who did not fix their own seeds. 

% \isabelle{It is not clear whether you changed the seed for these 3 runs.}
% \hung{I did not change the seeds. The performance of an agent only changes only if the participant did not fix their seeds. But maybe we should have changed the seeds...}

On each dataset $\mathcal{D}_i$, we evaluated an agent $\mathcal{A}_j$ by the Area under the Learning curve $ALC_i^j$ of the agent on the dataset. In the final ranking, the agent is ranked based on its average ALC over all datasets ($\mathcal{N}=30$ datasets):

\begin{equation}
    \mu_j =  \frac{\sum_{i=1}^{\mathcal{N}} ALC_i^j}{\mathcal{N}}
\end{equation}

To measure the variability of an agent $\mathcal{A}_j$, we computed the standard deviation of ALC scores obtained by the agent over all datasets:

\begin{equation}
    \sigma_j = \sqrt{\frac{\sum_{i=1}^{\mathcal{N}} (ALC_i^j - \mu_j)^2}{\mathcal{N}}}
\end{equation}

\section{Analyses of the first round results}
Results, final rankings, and prizes in the Final phase of the first round are shown in Table \ref{table:summary}. The 1st, 2nd, and 3rd ranked teams qualifying for prizes, as per the challenge rules\footnote{\firstcodalabpage}, were awarded prizes of 500\$, 300\$, and 200\$, respectively. In the remainder of this section, we give a more in-depth view of how each team performed on each dataset in this round, compared to our baselines.

In this round, the participants are asked to solve two tasks simultaneously: \textit{algorithm selection} and \textit{budget allocation}. Both of them are crucial to achieving our goal of maximizing the area under an agent's learning curve. We found approaches submitted by the participants using a wide range of methods, from simple (e.g. using algorithm ranking and pre-defined values for $\Delta t$) to more sophisticated (e.g. predicting scores and timestamps of unseen learning curve points). The results first indicate that using both learned policies (models) for choosing algorithms and spending time budget (used by team \textit{\MoRiHa} and \textit{\neptune}) yields better ALC scores than hard-coded ones (e.g. using a fixed pre-defined list of $\Delta t$ in \textit{\AIpert} and our \textit{DDQN} baseline).

According to Table \ref{table:result-details}, team \textit{\MoRiHa} obtained the highest average ALC of 0.43 in the final phase. 
% \lisheng{I think team jmhansel and \MoRiHa are the same team, and we should keep only one name to avoid confusing readers.}\hung{I fixed it.}
It succeeded in achieving the highest ALC score on 21 out of 30 datasets. In addition, it performed notably better than other teams in some datasets, such as \textit{tania}, \textit{robert}, \textit{newsgroups}, and \textit{marco}. They are datasets of either multi-label or multi-class classification tasks with a very high number of features. Moreover, most of them are sparse datasets, which is often seen as a challenge for learners. Further investigation will be done in our future work to understand the outperformance of \MoRiHa's method on these particular datasets. We will also encourage team \MoRiHa to join the second round to study the robutness of their results.
% \lisheng{Are we sure we are going to study this question? and in which occasion? This is an observation in the first round, I don't see where we can deliver the answer to this question.}
% \hung{One possibility is in a paper analyzing the results of both rounds.}
% \lisheng{OK. I think we can leave it this way, and in the 2nd round analysis paper, we can include the findings to the study of agents' behaviors in the 2 rounds}
% \isabelle{We can write: we will encourage team \MoRiHa to join the second round to study the robutness of their results.}
% \lisheng{Are these datasets similar in some sense? For example, they are all multi-class classification (10 - 95 classes); tania and marco are very sparse. The similarities can give insights that why \MoRiHa works so well on these datasets?}\hung{I added some texts for this.}
Team \textit{\neptune} has a score of 0.42 which is very close to the winner's, followed by team \textit{\AIpert} with a score of 0.40. Team \textit{\automlfreiburg}, which was ranked 4th, achieved a slightly lower score (0.37) than our DDQN baseline (0.38), and so did team \textit{\automlhannover} (0.32). 

The successes of the top-ranked team can be explained by the strategies they implemented. Team \textit{\MoRiHa}, which finished in the 1st place, uses a simple yet efficient approach that explores the most promising algorithms and avoids wasting time switching between too many different algorithms. Interestingly, team \textit{\neptune} learns a policy for allocating time budget using learning curve convergence speed. The Reinforcement Learning-based approach of team \textit{\AIpert} is very intuitive as our competition borrows the RL framework to formulate our problem, which also explains our baseline choice of DDQN. However, by complementing it with the K-means clustering method, \textit{\AIpert}'s approach achieved higher performance than our baseline. Both \textit{\AIpert} and \textit{\automlfreiburg} share the same idea of suggesting algorithms based on dataset similarity.

\begin{table*}[]
\centering
\caption{\textbf{Final phase ranking of the first round}. Teams are ranked based on their average ALC scores, which were recorded on the worst run among three runs. Team \textit{\MoRiHa} was ranked 1st, but not qualified for a monetary prize. Teams \textit{\neptune}, \textit{\AIpert}, and \textit{\automlfreiburg} are qualified for the prizes.}
\label{table:summary}
\begin{tabular}{|c|c|c|c|l|}
\hline
\rowcolor[HTML]{FFFFFF} 
\textbf{Rank} & \textbf{\begin{tabular}[c]{@{}c@{}}Team\\ (username)\end{tabular}} & \textbf{\begin{tabular}[c]{@{}c@{}}ALC \\ score\end{tabular}} & \textbf{\begin{tabular}[c]{@{}c@{}}Monetary Prize\\ qualification\end{tabular}} & \multicolumn{1}{c|}{\cellcolor[HTML]{FFFFFF}\textbf{\begin{tabular}[c]{@{}c@{}}Comments/\\source code\end{tabular}}}                                                                       \\ \hline
\rowcolor[HTML]{FFFFFF} 
1             & \begin{tabular}[c]{@{}c@{}}\textbf{\MoRiHa}\\ (username: jmhansel)\end{tabular}                                                  & 0.43                                                          &   NO            & \begin{tabular}[c]{@{}l@{}}This team is not qualified for\\ a monetary prize due to close\\ relation with the organizers\\ \firstteam \end{tabular} \\ \hline
\rowcolor[HTML]{FFFFFF} 
2             & \begin{tabular}[c]{@{}c@{}}\textbf{\neptune}\\ (username: \neptune)\end{tabular}                                                   & 0.42                                                          & OK         &  \begin{tabular}[c]{@{}l@{}} \secondteam \end{tabular}  \\ \hline
\rowcolor[HTML]{FFFFFF} 
3             & \begin{tabular}[c]{@{}c@{}}\textbf{\AIpert}\\ (username: \AIpert)\end{tabular}                                                     & 0.40                                                          & OK         &   \thirdteam                                                                                                                                   \\ \hline

\rowcolor[HTML]{FFFFFF} 
4             & \begin{tabular}[c]{@{}c@{}}\textbf{\automlfreiburg}\\(username: \automlfreiburg)\end{tabular}                                           & 0.37                                                          & OK          &   \fourthteam                                                                                                                                  \\ \hline
\rowcolor[HTML]{FFFFFF} 
5             & \begin{tabular}[c]{@{}c@{}}\textbf{\automlhannover} \\ (username: \amsks)\end{tabular}                                                     & 0.32                                                          &                & \fifthteam                                                                                                                                      \\ \hline
\rowcolor[HTML]{FFFFFF} 
6             &\begin{tabular}[c]{@{}c@{}}\textbf{\pprp} \\ (username: \pprp) \end{tabular}                                                         & 0.24                                                          &                &                                                                                                                                      \\ \hline
\rowcolor[HTML]{FFFFFF} 
7             & \begin{tabular}[c]{@{}c@{}}\textbf{\arushsharma} \\ (username: \arushsharma) \end{tabular}                                                    & 0.23                                                          &                &                                                                                                                                      \\ \hline
\rowcolor[HTML]{FFFFFF} 
8             & \begin{tabular}[c]{@{}c@{}}\textbf{\Xavier} \\ (username: \Xavier) \end{tabular}                                                           & 0.17                                                          &                &                                                                                                                                      \\ \hline
\end{tabular}
\end{table*}

%=================================

% \usepackage{multirow}
% \usepackage{colortbl}

% \begin{landscape}

% \begin{sidewaystable*}
% \usepackage{multirow}
% \usepackage{colortbl}

% \usepackage{multirow}
% \usepackage{colortbl}

% \usepackage{multirow}
% \usepackage{colortbl}

% \usepackage{multirow}
% \usepackage{colortbl}

% \usepackage{multirow}
% \usepackage{colortbl}

% \usepackage{multirow}
% \usepackage{colortbl}

% \usepackage{multirow}
% \usepackage{colortbl}

% \usepackage{multirow}
% \usepackage{colortbl}

\renewcommand{\arraystretch}{1.2}

\begin{table*}
\centering
\caption{\textbf{ALC scores of the top 5 methods: \MoRiHa~(AT01), \neptune~(AT02), \AIpert~(AT03), \automlfreiburg~(AT04), \automlhannover~(AT05), and our baselines: Double Deep Q Network (DDQN), Best on Samples (BOS), Freeze-Thaw BO (FT), Average Rank (AR), Random Search (RS)}. The reported scores correspond to the worst of 3 runs for each method. The last row shows the average ALC scores (in descending order, from left to right) over 30 datasets.}
% \isabelle{Anonymize Winners}
% \isabelle{You do not want to put your nice heatmap?}
% \hung{I put the heatmap below.}
\label{table:result-details}
\includegraphics[width=0.9\textwidth]{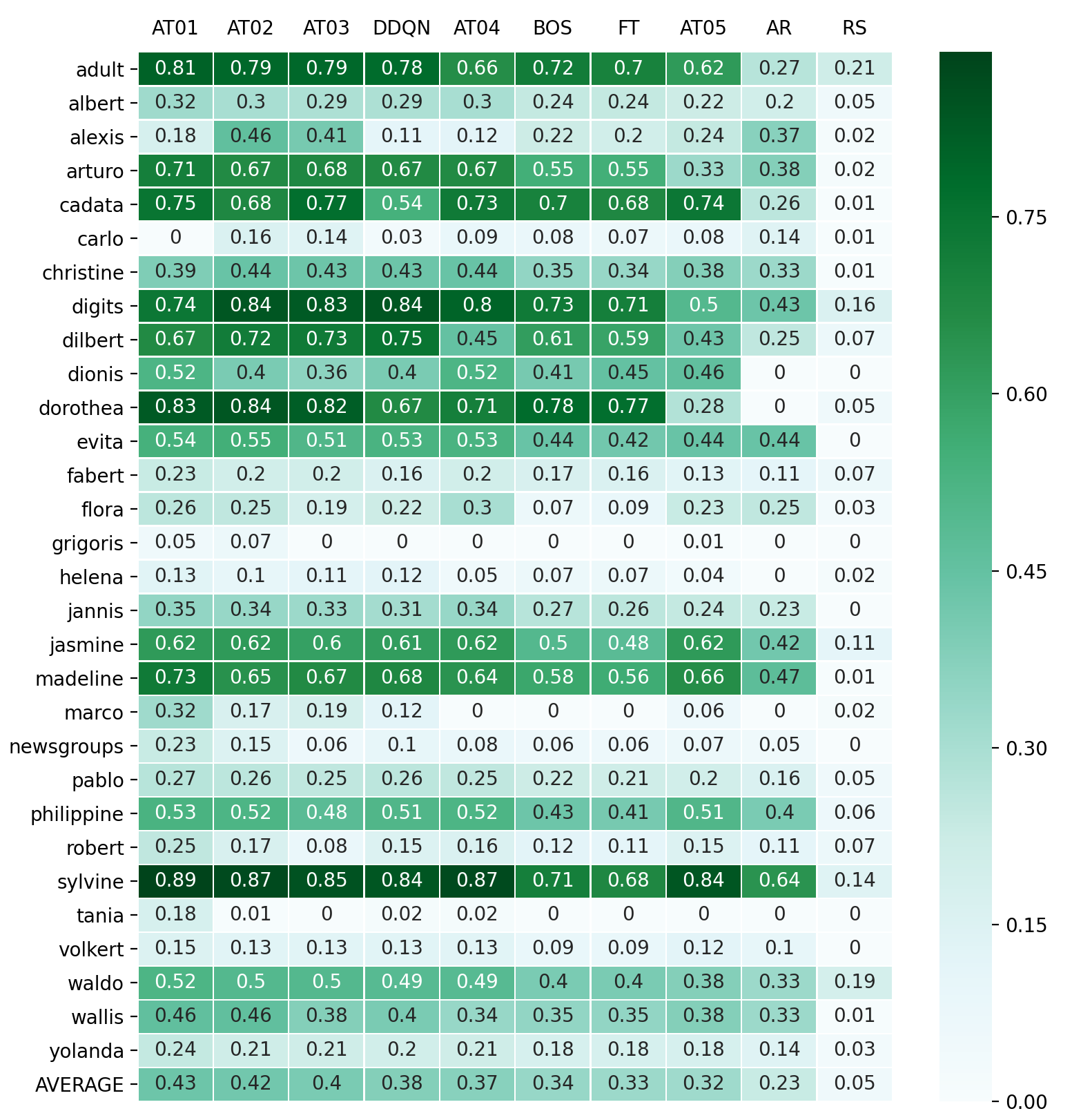}
\end{table*}

\section{Winning solutions}
In this section, we briefly introduce strategies of the winning solutions. More details on their implementations can be found in our factsheet summary (Appendix \ref{sec:factsheet}) or in individual factsheets provided by the winning teams in Table \ref{table:summary}.

\subsection{Team \textit{\MoRiHa} (1st place)}
According to team \textit{\MoRiHa}, they focus on ``doing the right thing at the right time". Their agent is very goal-oriented (i.e. maximizing the area under the learning curve) and does not rely on complex models or expensive computations. They emphasize the importance of having an accurate schedule regarding the invested time (choosing $\Delta t$ for each query) in order to avoid wasting time budget. One of their key findings is that switching between algorithms during exploration is very costly and it is rarely beneficial to switch the explored algorithm more than once. They build a \textbf{time model} for each algorithm to predict the time when the \textbf{first point} on the algorithm's learning curve is available. In addition, they keep a \textbf{list of algorithms ranked descending} based on their ALC in the meta-training phase. In meta-testing their algorithm scheduler explores algorithms in an order starting from the best algorithm. If an algorithm's learning curve stales, the next algorithm will be chosen. Time allocation is done using the time models and a \textbf{heuristic} procedure. This is the only team having an assumption that the learning curves are monotonically increasing.

\subsection{Team \textit{\neptune} (2nd place)}
As described by team \textit{\neptune}, they train a learning curve predictor to \textbf{predict unseen points} on the learning curves (i.e. by interpolating the original scores) in order to find the best algorithm for a new dataset. In addition, they train an algorithm classifier to categorize algorithms into three groups based on their \textbf{learning curve convergence speed}: Fast/Median/Slow. Different budget allocation strategies will be selected according to the algorithm's convergence type. Regarding their implementation, they train MLP networks to perform both tasks: learning curve prediction and algorithm classification.

\subsection{Team \textit{\AIpert} (3rd place)}
According to team \textit{\AIpert}, their method aims at uncovering good algorithms as fast as possible, using a low-computational cost and simple process. The novelty lies in the combination of an off-policy Reinforcement Learning method (\textbf{Q-learning}) and \textbf{K-means clustering} model. As similar datasets usually have the same \textit{learning behavior}, organizing similar datasets into groups based on their meta-features is essential. They thus build a Q-learning matrix for each dataset cluster (12 clusters in total). In meta-training, each dataset is seen multiple times and the corresponding Q-learning matrix is updated. This allows the agents to be exposed to more situations (different observations and rewards) on a dataset. In meta-testing, they first determine which cluster the given dataset belongs to. Then, the Q-learning matrix associated with the cluster is utilized as a policy to guide the agent. $\Delta t$ is not chosen as an absolute value but from \textbf{a fixed list of time portions} of the total time budget.

\subsection{Team \textit{\automlfreiburg} (4th place)}
As described by team \textit{\automlfreiburg}, their algorithm selection policy is based on a \textbf{Deep Gaussian Processes} (DGP) surrogate, in a similar way to FSBO \cite{fsbo}. The surrogate aims to predict the performance of an algorithm at time $t + \Delta t$, with $\Delta t$ chosen by a \textbf{trained budget predictor}. The DGP is trained during the meta-training phase, and fine-tuned in the meta-testing phase. During meta-training, they store the best algorithm for each dataset to be used as the first algorithm to query during meta-testing. The ``best" algorithm is defined as the one that has the highest $\frac{y_0}{t_0}$, which means they favor algorithms that achieve \textbf{high performance early}. Given a new dataset in meta-testing, the best algorithm of the \textbf{closest dataset} (previously seen in meta-training, based on the euclidean
distance) will be selected. During meta-testing, they keep updating the observed algorithms and fine-tune the DGP surrogate.

\section{Lessons learned and new design}
In this section, we describe the limitations of our original design and how we addressed them in the second round.

First, one set of limitations arises because we precomputed learning curves (performance as a function of time) and therefore have a fixed predetermined sampled time points. In our first challenge edition, we opted to interpolate in between time points with the last recorded performance. The criticism of participants was that in real life, if they chose an in between time point, they would get newer information. To mitigate that, in the new challenge round, the participants cannot choose any time point (thus no need to interpolate), they have to choose the next pre-computed learning curve point.

% First, the time when a new point on a learning curve appears quite randomly to the participants, making it difficult to predict and learn how to spend the given time budget wisely. Second, if the time budget $\Delta t$ allocated by an agent is not enough to query a new point on a model's learning curve, $\Delta t$ will be subtracted from the total time budget $\mathcal{T}$ without returning any new learning curve information. This may not be realistic because in a real-life scenario, if one decided to train a model in $\Delta t$, a new model should be returned along with a new accuracy score.

% \isabelle{The previous sentence is unclear.}
% \hung{I elaborated it.}

% \begin{figure*}
%     \centering
%     \includesvg[width=1\textwidth]{fig/digits_test.svg}
%     \caption{Caption}
%     \label{fig:my_label}
% \end{figure*}

We thus provide a new type of learning curve for the second round: learning curve as a function of training data size, as opposed to learning curves as a function of time. The time budget for querying a point on a learning curve will be returned by the environment and not chosen by the agent, which means that the agent has to pay whatever it costs.

Second, the test learning curves were highly correlated with the validation curves. Therefore, one could overfit the former by simply overfitting the latter. In the second round, the agent will always be evaluated using the test learning curves but on a completely different set of datasets in each phase (Development phase and Final phase).

The second round of our challenge was accepted at the \automl. Participation in the first round is not a prerequisite for the second round. The second round comes with some \textbf{new features}, including:

\begin{itemize}
    \item \textbf{Learning curve}: we focus on learning curves as functions of training data size. We thus collected a new large meta-dataset of such learning curves.
% the max features hyperparameter
    \item \textbf{Competition protocol}: Given a portfolio of algorithms, an agent suggests which algorithm and the amount of training data to evaluate the algorithm on a new task (dataset) efficiently. The agent observes information on both the \textit{training learning curves} and \textit{validation learning curves} to plan for the next step. Test learning curves, which are kept hidden, will be used for evaluating the agent. 
    \item \textbf{Data split}: We use half of the meta-dataset for the Development phase and the other “fresh” half to evaluate the agent in the Final phase.
\end{itemize}

The second round is split into three phases:
\begin{itemize}
    \item \textbf{Public phase (1 week)}: participants practice with the given starting kit and sample data
    \item \textbf{Development phase (6 weeks)}: participants submit agents that are meta-trained and meta-tested on the platform. 15 datasets will be used in this phase. 
    \item \textbf{Final phase (1 week)}: no further submissions are made in this phase. Your last submission in the Development phase will be forwarded automatically to this phase and evaluated on 15 fresh datasets (not used in the Development phase). 
\end{itemize}

Like in the first round, this is a competition with code submission and the participants do not see the data in either phase (only their submitted agent is exposed to the meta-datasets). 

We created a new meta-dataset of pre-computed learning curves of 40 algorithms with different hyperparameters on 30 datasets used in the AutoML challenge. The algorithms are created from four base algorithms: K-Nearest Neighbors (KNN), Multilayer Perceptron (MLP), Adaboost, Stochastic Gradient Descent (SGD), but with different values of hyperparameters. We added meta-features of datasets and hyperparameters of algorithms. We respected the data split of the AutoML challenge to produce three sets of learning curves for each task, from the \textbf{training, validation, and test sets}. The type of metric used to compute the learning curves of the meta-dataset is provided in the meta-features of the dataset. We also generated a new synthetic meta-dataset that contains 12000 learning curves, in a similar way used in the first round, but with a new type of learning curve as explained above (learning curve as a function of training data size).

% \isabelle{In the next paragraph, explain what is new/different compared to the first round.}
% \hung{I think I have explained the differences on the previous page.}
% \lisheng{So we no longer use the artificial meta-dataset?}
% \hung{I added some texts for this.}
In meta-training, the following data is given to the agent to meta-learn: meta-features of datasets, hyperparameters of algorithms, training learning curves, validation learning curves, and test learning curves. While in meta-testing, the agent interacts with an environment in a Reinforcement Learning style. Given a portfolio of algorithms, an agent suggests which algorithm and the amount of training data to evaluate the algorithm on a new task (dataset) efficiently. The agent observes information on both the training learning curve and validation learning curve to plan for the next step. An episode ends when the given time budget is exhausted. The following two lines of code demonstrate the interactions between the agent and the environment:

\begin{center}
    $action = trained\_agent.suggest(observation)$

$observation, done = env.reveal(action)$
\end{center}

with:

\textbf{observation} : a tuple of\\ $(A, p, t, R\_train\_A\_p, R\_validation\_A\_p)$, with:

\begin{itemize}
    \item $A$: index of the algorithm provided in the previous action,
    \item $p$: decimal fraction of training data used, with the value of p in [0.1, 0.2, 0.3, ..., 1.0]
    \item $t$: the amount of time it took to train A with training data size of p, and make predictions on the training/validation/test sets.
    \item $R\_train\_A\_p$: performance score on the training set
    \item $R\_validation\_A\_p$: performance score on the validation set
\end{itemize}

\textbf{action}: a tuple of $(A, p)$, with:
\begin{itemize}
    \item $A$: index of the algorithm to be trained and tested
    \item $p$: decimal fraction of training data used, with the value of p in [0.1, 0.2, 0.3, ..., 1.0]
\end{itemize}

The scoring program automatically chooses the best algorithm at each time step (i.e. the algorithm with the highest validation score found so far, which is different from the first round where it was chosen by the agent) to compute the agent’s test learning curve (as a function of time spent). The metric used for ranking on the leaderboard is the Area under the agent's Learning Curve (ALC).
% \lisheng{Do we rank participants in Development phase too? Because we have a meta-testing in this phase.}
% \hung{We do rank them just to give them an idea of how good their solutions are compared to others. However, the ranking in Development phase is not taken into account for giving prizes.}

\subsection{Baseline results of the second round}
% \isabelle{There is a section missing here with baseline results for the second round.}
We re-use our baselines from the first round, including: Double Deep Q Network, Best on Samples, Freeze-Thaw Baysian Optimization, Average Rank, and Random Search agents. Their implementations are taken from the first round (\textit{Anonymous-paper}) with some modifications in order to work with our new protocol and meta-dataset. We also provide them (except the Random Search agent) a smarter policy for choosing the training data size in each step. Each agent keeps track of the algorithms ($A$) and the training data size ($p$) tested on the algorithms. Every time it re-selects an algorithm, it should increase $p$ by 0.1 to go forward on the algorithm learning curve. 

We run each method three times and report the worst run of each of them in Table \ref{table:2nd-round-baselines}. Similar to the first round, among the baselines, \textbf{DDQN} still performs best in the second round with an average ALC of 0.38. It is the winner (co-winner) of 16 out of 30 datasets with the highest performance difference compared to other baseline methods seen on the dataset \textit{tania}. \textbf{BOS} and \textbf{FT} are not far behind with average scores of 0.37 and 0.36, respectively. The improvement of \textbf{BOS} compared to the first round can be explained by the adaptation of its strategy in this round. It tries each method on a small subset of training data first (e.g. 10 percent of training samples), instead of spending a fixed amount of time as in the previous round. This should help it not waste a time budget since choosing an adequate amount of time to get a new point on learning is no longer necessary in the second round. The success of \textbf{BOS} suggests that information on the first point of the learning curve is crucial to decisions on selecting algorithms.  

\textbf{AR} and \textbf{RS} perform poorly with the same score of 0.28. Although focusing only on one algorithm as what \textbf{AR} method does can bring benefits in some cases (e.g. on dataset \textit{cadata} and \textit{didonis}), a dataset-dependent policy for selecting algorithms is still necessary to be successful on multiple cross-domain tasks.

\begin{table*}
\centering
\caption{\textbf{Baseline results in the second round.} We show ALC scores of Double Deep Q Network (DDQN), Best on Samples (BOS), Freeze-Thaw Bayesian Optimization (FT), Average Rank (AR), and Random Search (RS) on each dataset, in descending order of the average ALC scores (on the last row), from left to right. The first 15 datasets are used in the Development phase, the rest is for the Final phase.}
\label{table:2nd-round-baselines}
\includegraphics[width=0.7\textwidth]{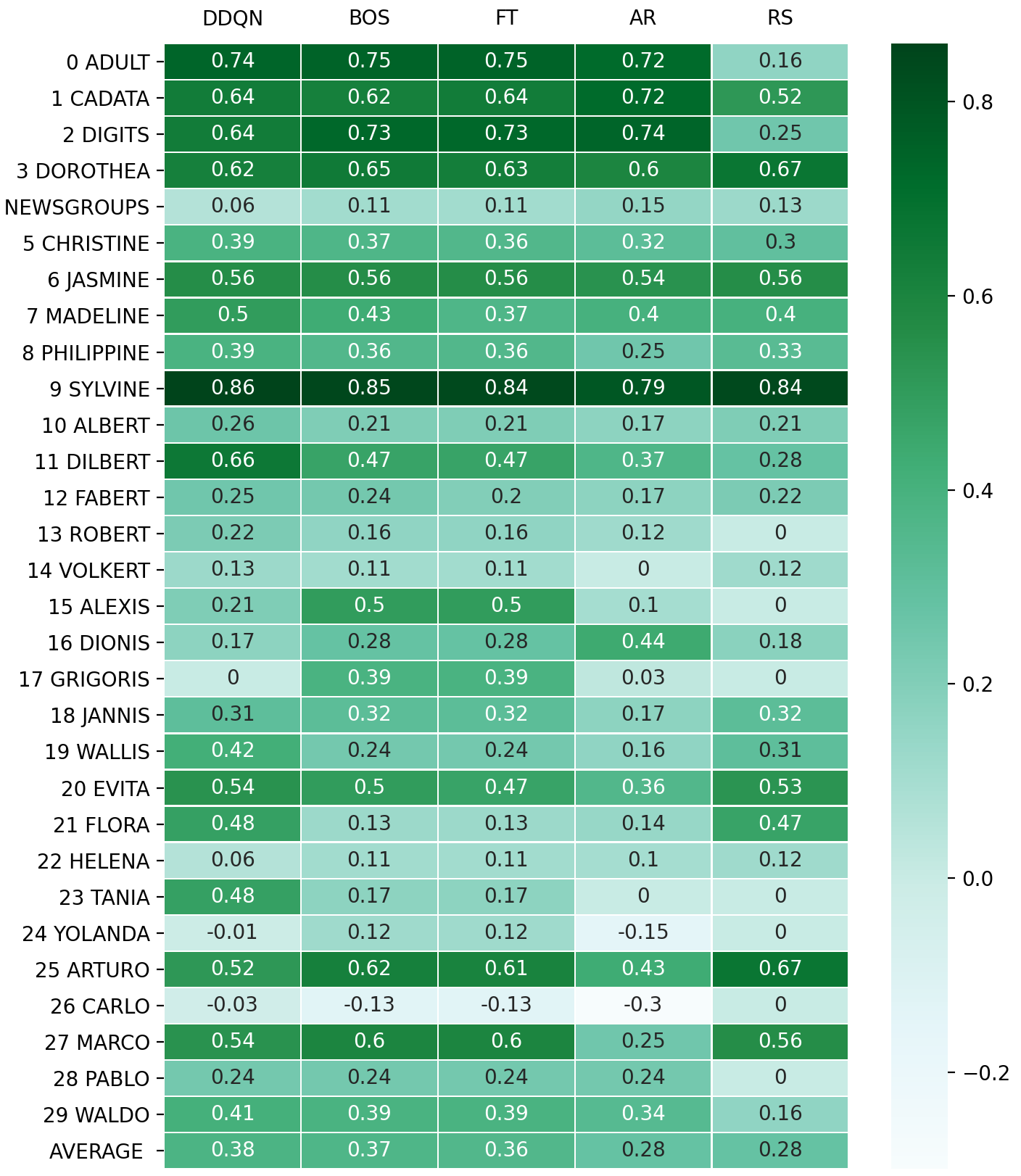}
\end{table*}

\section{Conclusion}

The first round results of our challenge have revealed that agents that learn both
policies for selecting algorithms and allocating time budget are more successful in our challenge setting. Team \textit{\MoRiHa}, who finished in 1st place, outperformed all other teams and our baselines on two-thirds of the datasets. We propose a novel setting and a new meta-dataset for the second round and present baseline results. \textbf{DDQN} maintains its advantages in the second round, while an intuitive and simple baseline as \textbf{BOS} can work quite well in the second round. We are looking forward to see whether the findings of the first round will be reinforced and whether participants' solutions can outperform our baselines significantly in the second round.

For our future work, we want to perform more post-challenge analyses to verify whether progress was made in meta-learning from learning curves. First, we would like to do a point-by-point comparison of the winning methods in the first round, based on their fact sheets. Second, to investigate further the winning methods and see what contributed the most to their success, we want to perform systematic experiments in collaboration with the winners. More concretely, we will build a common workflow and ask participants to conduct ablation studies. Lastly, we are also interested in examining
the effect of changes in our reward function hyper-parameters on participants' performances. 

%the effect of the hyperparameter $t_0$ (part of the reward function used in our scoring program) on the participants' performances. 
\section*{Acknowledgment}
We would like to thank challenge participants for providing feedback and sharing their methods. We are grateful to Chalearn for donating prizes and Google for providing computing resources. This work was supported by ChaLearn, the ANR Chair of Artificial Intelligence HUMANIA ANR-19-CHIA-0022 and TAILOR EU Horizon 2020 grant 952215.
 
\section*{Software and Data}

All software (including starting kit and winning solutions) is open-sourced on our website (\metalearningwebsite). The meta-datasets will remained private on the challenge platform (\codalab) to serve as a long-lasting benchmark for research in meta-learning.
% \isabelle{Anonymize Codalab}
% \isabelle{Complete this section}
% \hung{I added some texts.}

% ANONIMIZATION??

% \section*{Acknowledgements}
% (anonymized)
% EMPTY?

\bibliographystyle{icml2022}
\newpage
\bibliography{refs}

%%%%%%%%%%%%%%%%%%%%%%%%%%%%%%%%%%%%%%%%%%%%%%%%%%%%%%%%%%%%%%%%%%%%%%%%%%%%%%%
%%%%%%%%%%%%%%%%%%%%%%%%%%%%%%%%%%%%%%%%%%%%%%%%%%%%%%%%%%%%%%%%%%%%%%%%%%%%%%%
% APPENDIX
%%%%%%%%%%%%%%%%%%%%%%%%%%%%%%%%%%%%%%%%%%%%%%%%%%%%%%%%%%%%%%%%%%%%%%%%%%%%%%%
%%%%%%%%%%%%%%%%%%%%%%%%%%%%%%%%%%%%%%%%%%%%%%%%%%%%%%%%%%%%%%%%%%%%%%%%%%%%%%%
\newpage
\appendix
\onecolumn

% You can have as much text here as you want. The main body must be at most $8$ pages long.
% For the final version, one more page can be added.
% If you want, you can use an appendix like this one, even using the one-column format.
% %%%%%%%%%%%%%%%%%%%%%%%%%%%%%%%%%%%%%%%%%%%%%%%%%%%%%%%%%%%%%%%%%%%%%%%%%%%%%%%
% %%%%%%%%%%%%%%%%%%%%%%%%%%%%%%%%%%%%%%%%%%%%%%%%%%%%%%%%%%%%%%%%%%%%%%%%%%%%%%%
\section{Factsheet Summary}
\label{sec:factsheet}
In this section, we summarize the information provided in the participants' factsheets. Only top 5 teams submitted their factsheets, including: \textit{\MoRiHa, \neptune, \AIpert, \automlfreiburg,} and \textit{\automlhannover.}
\subsection{PRE-PROCESSING \& FEATURE ENGINEERING}
\textbf{\ul{Question 1: Did you perform any data pre-processing methods?}}
\begin{figure}[H]
    \centering
    \includegraphics[width=0.8\textwidth]{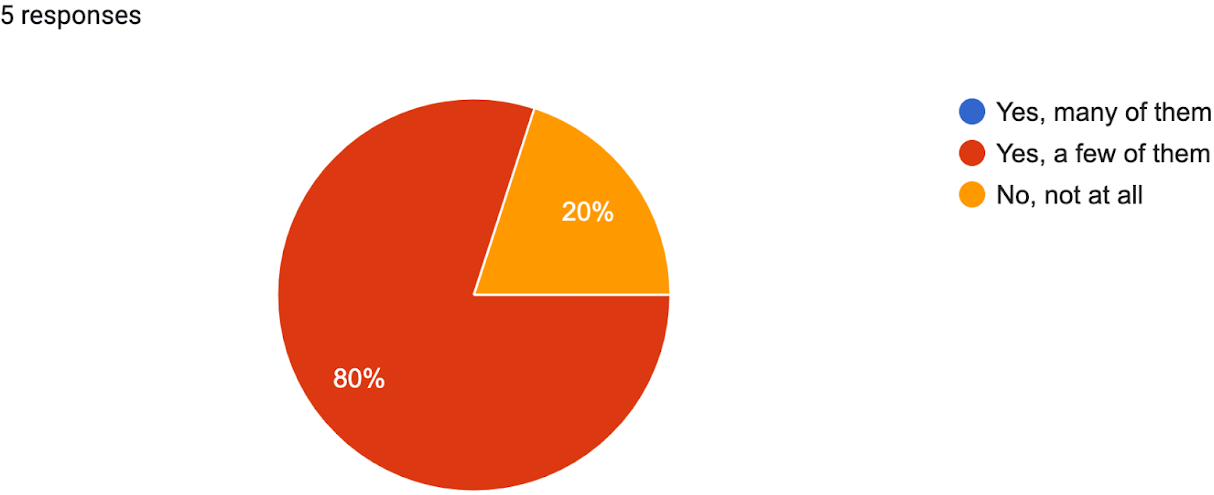}
    \caption{Pre-processing methods include: Standard Scaler, Feature Scaling, One-hot Encoding, extracting the 90\% convergence time and final performances.}
    \label{fig:my_label}
\end{figure}

\textbf{\ul{Question 2: Did you perform feature engineering methods?}}
\begin{figure}[H]
    \centering
    \includegraphics[width=0.8\textwidth]{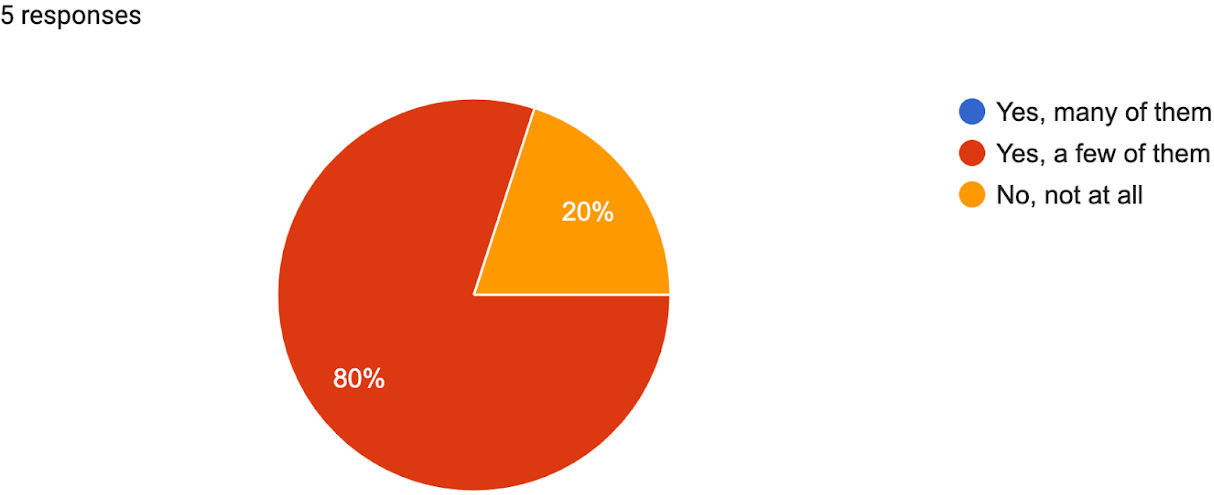}
    \caption{Feature engineering methods include: Clustering; Polynomial features of train\_num, feat\_num; Multi-step forecasting for data augmentation.}
    \label{fig:my_label}
\end{figure}

\subsection{DATA USED FOR LEARNING}

\textbf{\ul{Question 3: Did you use all points on the learning curves or only some of them?}}
\begin{figure}[H]
    \centering
    \includegraphics[width=0.9\textwidth]{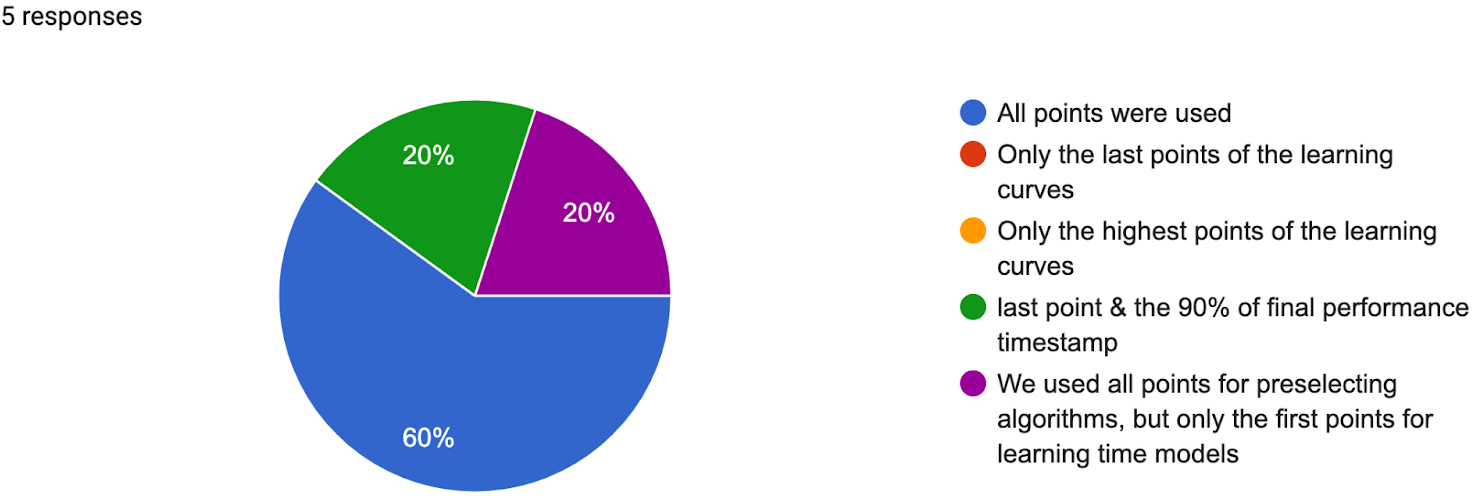}
    \caption{Most of the methods use all points on the learning curves for learning.}
    \label{fig:my_label}
\end{figure}

\textbf{\ul{Question 4: Did you make use of meta-features of datasets?}}
\begin{figure}[H]
    \centering
    \includegraphics[width=0.65\textwidth]{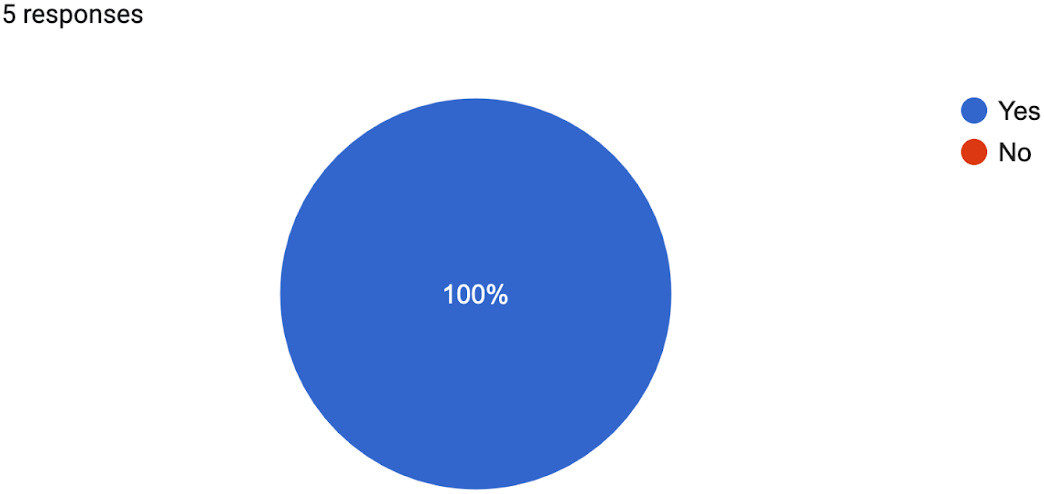}
    \caption{All methods took advantage of meta-features of datasets.}
    \label{fig:my_label}
\end{figure}

\textbf{\ul{Question 5: Did you implement a Hyperparameter Optimization component for your agent using the provided hyperparameters of algorithms?}}
\begin{figure}[H]
    \centering
    \includegraphics[width=0.65\textwidth]{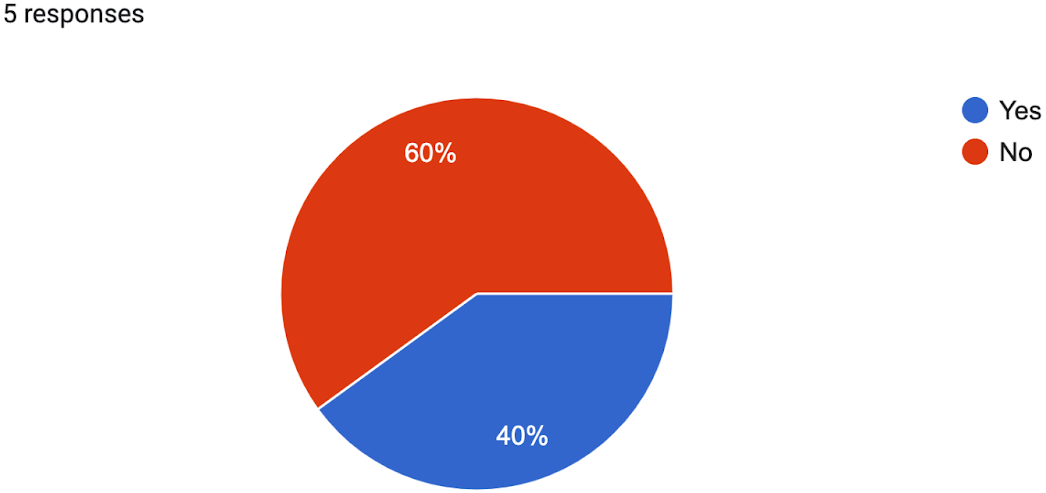}
    \caption{Some HPO tools were used, such as Hyperband
and FSBO.}
    \label{fig:my_label}
\end{figure}

\textbf{\ul{Question 6: In case you used either or both meta-features of datasets and algorithms, did it improve the performance of your method?}}
\begin{figure}[H]
    \centering
    \includegraphics[width=1\textwidth]{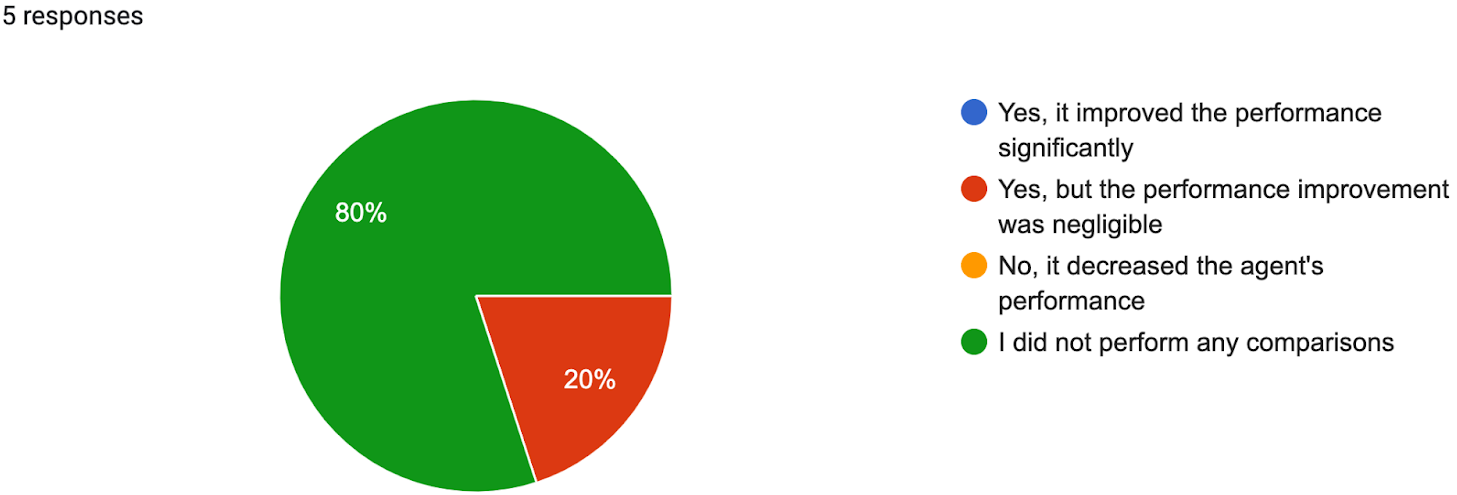}
    \caption{More experiments need to be done to confirm whether meta-features of datasets and algorithms help.}
    \label{fig:my_label}
\end{figure}

\textbf{\ul{Question 7: In any case, did you find the meta-features useful in our meta-learning setting?}}
\begin{figure}[H]
    \centering
    \includegraphics[width=1\textwidth]{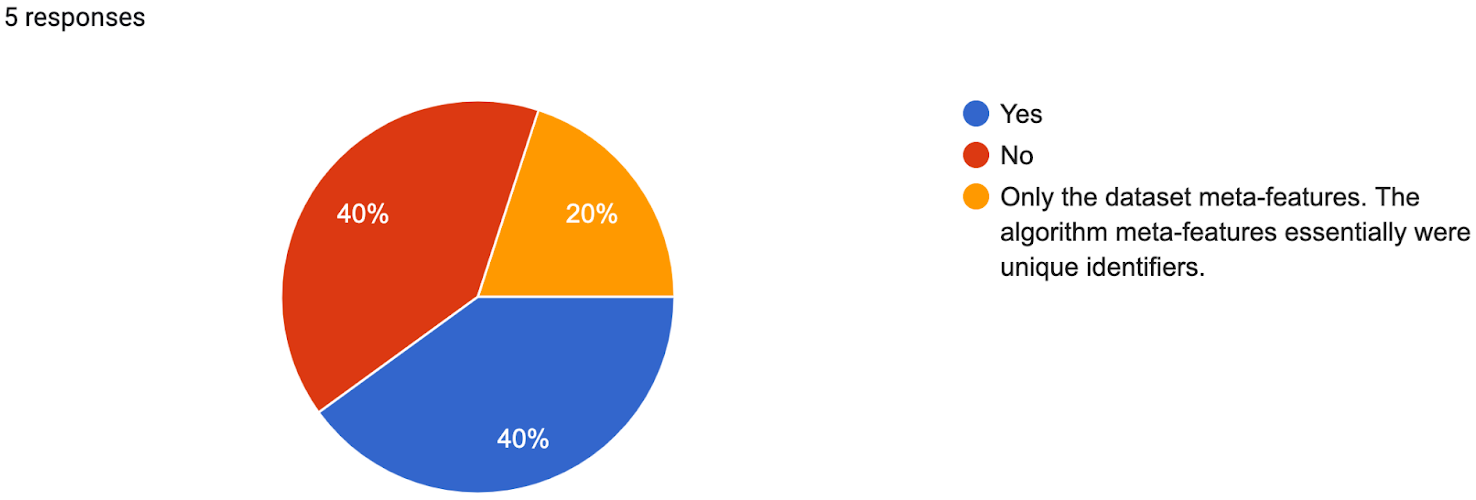}
    \caption{Not all participants find the provided meta-features useful.}
    \label{fig:my_label}
\end{figure}

\subsection{POLICY CHARACTERISTICS}

\textbf{\ul{Question 8: Does your agent learn a policy from datasets in the meta-training phase?}}
\begin{figure}[H]
    \centering
    \includegraphics[width=0.7\textwidth]{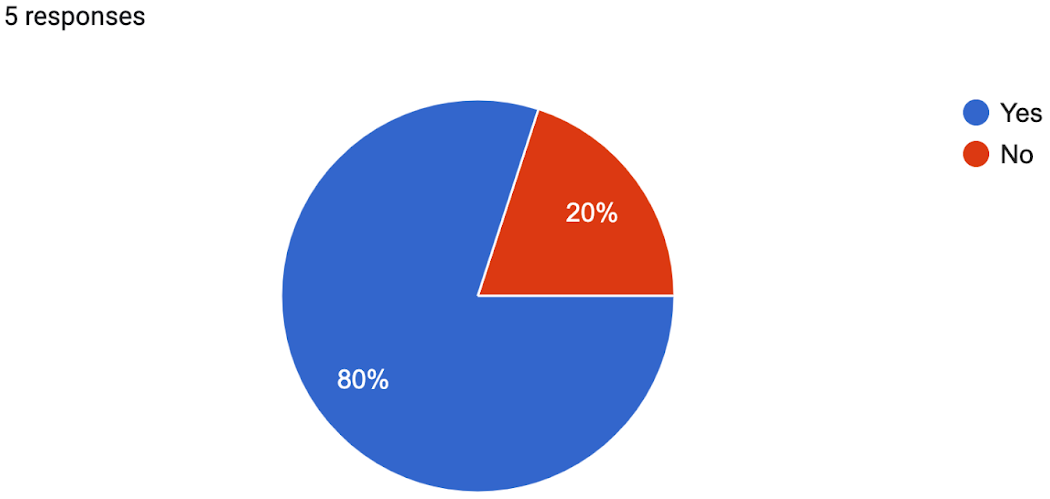}
    \caption{Most of the agents use a learned policy from the meta-training phase.}
    \label{fig:my_label}
\end{figure}

\ul{\textbf{Question 9: How does your agent manage the exploration-exploitation trade-offs (revealing a known good algorithm's learning curve vs. revealing a new algorithm candidate's learning curve )?}}

\begin{enumerate}
    \item With an $\epsilon$ greedy policy, in a Reinforcement Learning framework, only in meta- training we create different Q-matrices. In the meta-testing phase, we perform the choice of the new algorithm with the computed Q-matrices.
    
    \item We are very restrictive with switching the explored algorrithm. We preselect the single best performing algorithm from the validation learning curves statically and only explore other algorithms, if its learning curve is stale. So we strongly emphasize exploiting the single best algorithm.
    
    \item A modified Round Robin on the top-k performing algorithms. The incumbent i.e. the value of the top performing algorithm on the test dataset is challenged by zero budget allocated algorithms. Since the training on the validation data allows for immediate look up in the test dataset.

    \item Bayesian Optimization

    \item We find the best algorithm by a learning curve predictor.
\end{enumerate}

\ul{\textbf{Question 10: Does your agent switch between learning curves during an episode (i.e. switching between training different algorithms on the dataset at hand)?}}
\begin{figure}[H]
    \centering
    \includegraphics[width=0.9\textwidth]{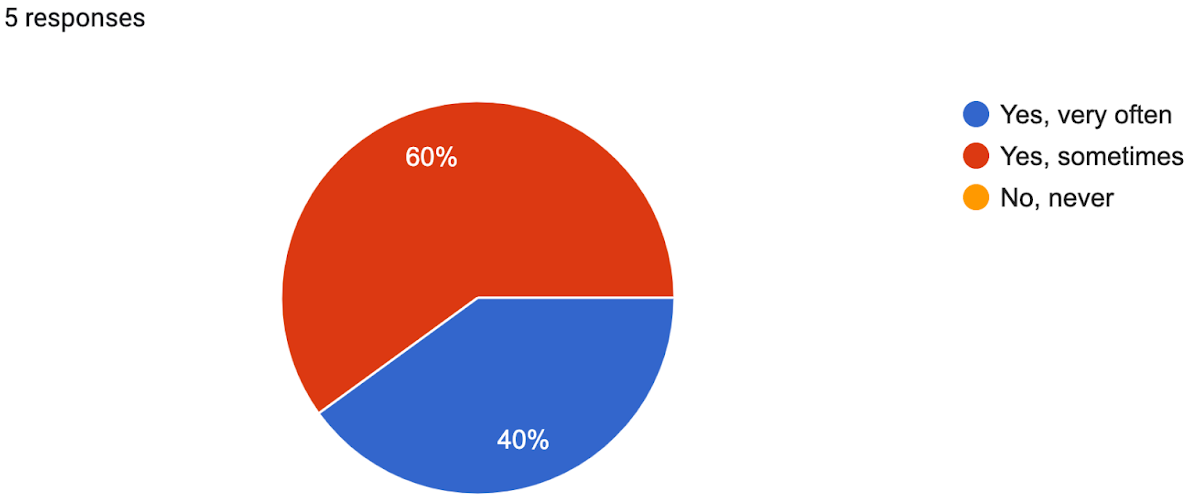}
    \caption{All agents switch between learning curves to find the best algorithm for the task at hand.}
    \label{fig:my_label}
\end{figure}

\ul{\textbf{Question 11: Does your agent leverage partially revealed learning curves on the dataset at hand?}}
\begin{figure}[H]
    \centering
    \includegraphics[width=0.8\textwidth]{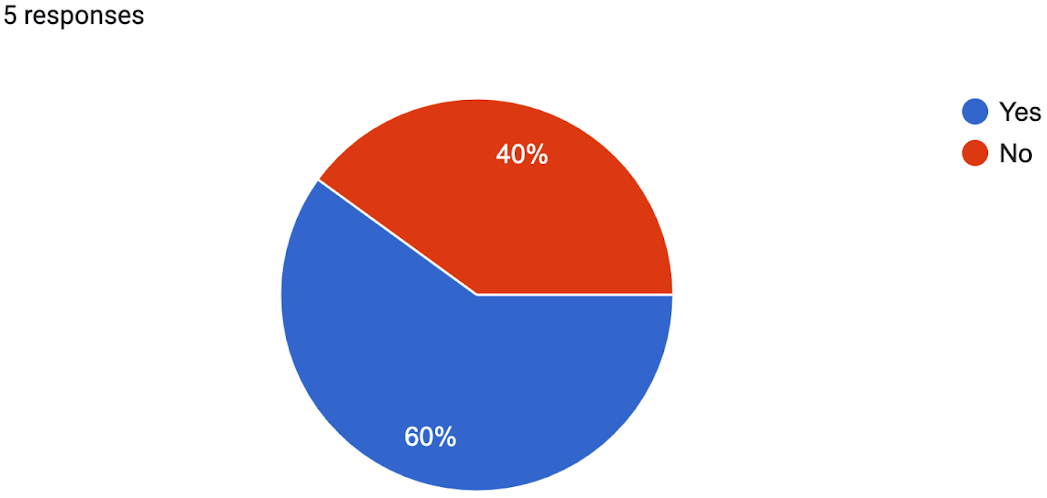}
    \caption{Some agents do not take into account information of partial learning curves on the task at hand for deciding their actions.}
    \label{fig:my_label}
\end{figure}

\ul{\textbf{Question 12: Did you make any assumptions about the shapes of the learning curves?}}
\begin{figure}[H]
    \centering
    \includegraphics[width=0.7\textwidth]{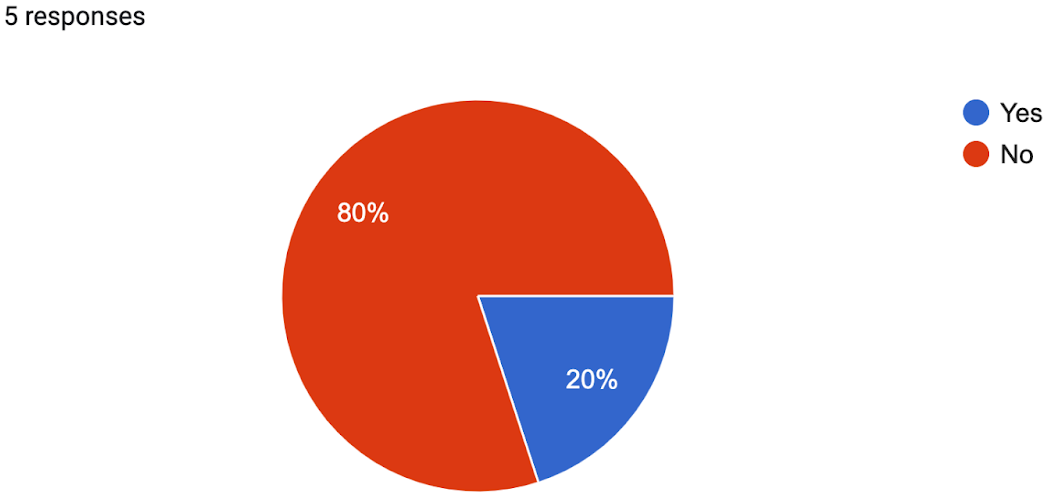}
    \caption{Only one team made an assumption that the learning curves are \textit{monotonically increasing}.}
    \label{fig:my_label}
\end{figure}

\ul{\textbf{Question 13: Does your agent predict unseen points on a learning curve?}}
\begin{figure}[H]
    \centering
    \includegraphics[width=0.7\textwidth]{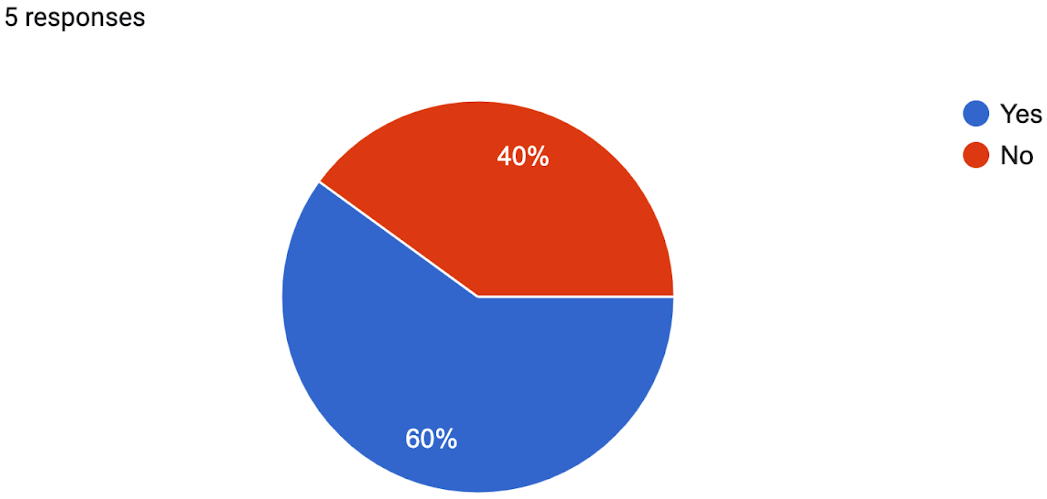}
    \caption{3 out of 5 agents make decisions based on predicted performance scores.}
    \label{fig:my_label}
\end{figure}

\ul{\textbf{Question 14: Does your agent perform pairwise comparisons of algorithms' learning curves?}}
\begin{figure}[H]
    \centering
    \includegraphics[width=1\textwidth]{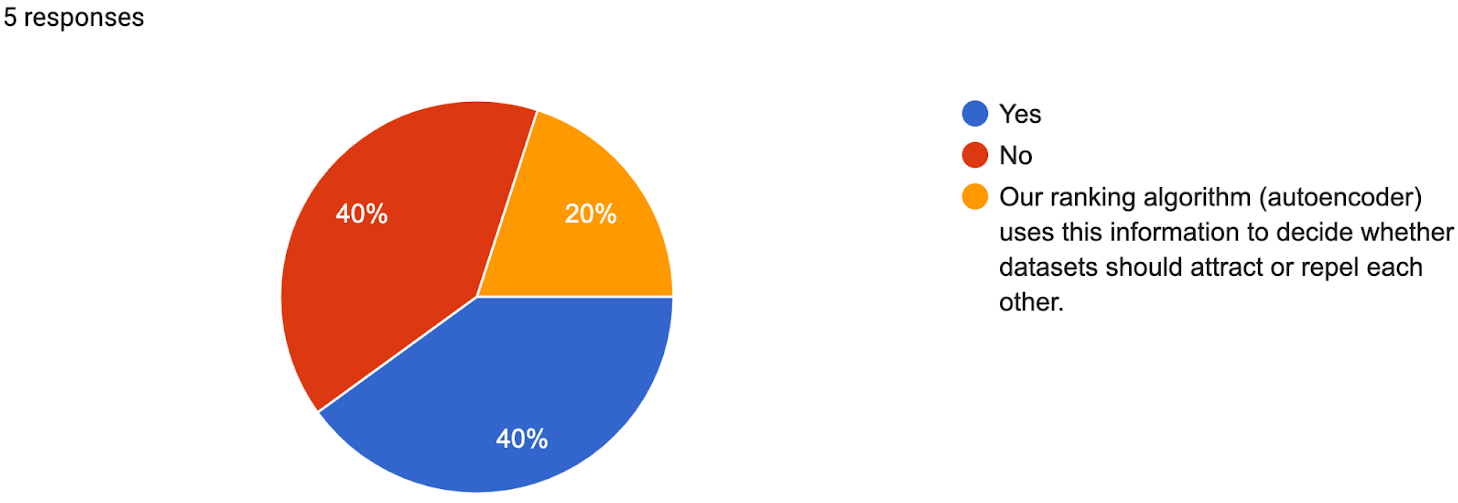}
    \caption{Pairwise comparisons of algorithms' learning curves have been exploited in 60\% of the agents.}
    \label{fig:my_label}
\end{figure}

\ul{\textbf{Question 15: How does your agent spend the given time budgets (i.e. choosing delta\_t at each step)?}}
\begin{figure}[H]
    \centering
    \includegraphics[width=1\textwidth]{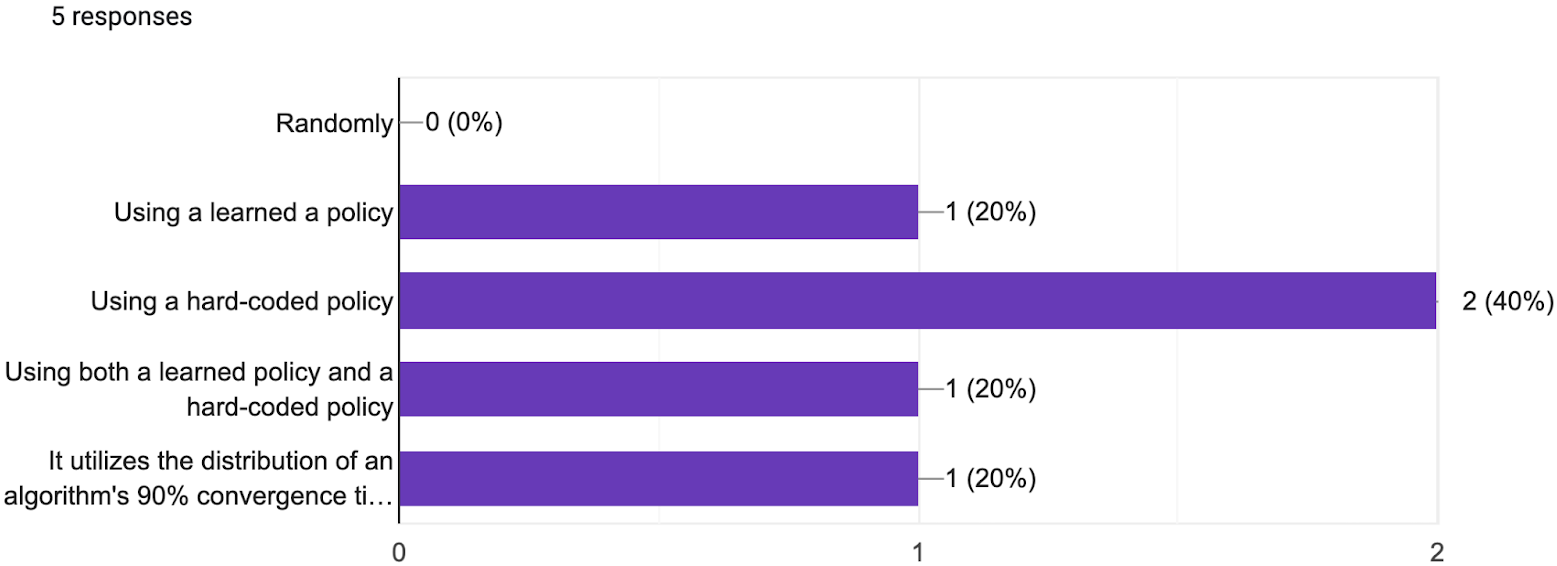}
    \caption{Participants use either or both hard-coded policy and learned policy to distribute a given time budget (no one does it randomly).}
    \label{fig:my_label}
\end{figure}

\ul{\textbf{Question 16: How does your agent choose which algorithm to contribute to its learning curve at each step (i.e. choosing A\_star)?}}
\begin{figure}[H]
    \centering
    \includegraphics[width=1\textwidth]{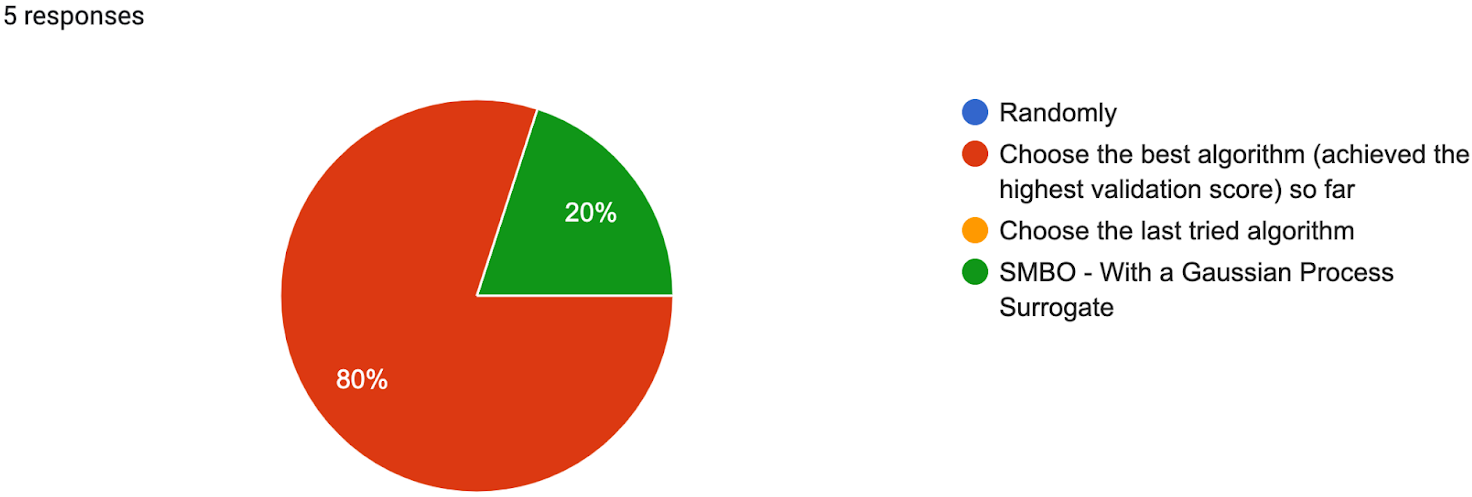}
    \caption{Most of the agents choose the best algorithm so far as $A^*$, with only one exception that uses Sequential Model Based Optimisation (SMBO) with a Gaussian Process Surrogate.}
    \label{fig:my_label}
\end{figure}

\ul{\textbf{Question 17: Which phase did you focus on more to improve your agent's performance?}}
\begin{figure}[H]
    \centering
    \includegraphics[width=0.9\textwidth]{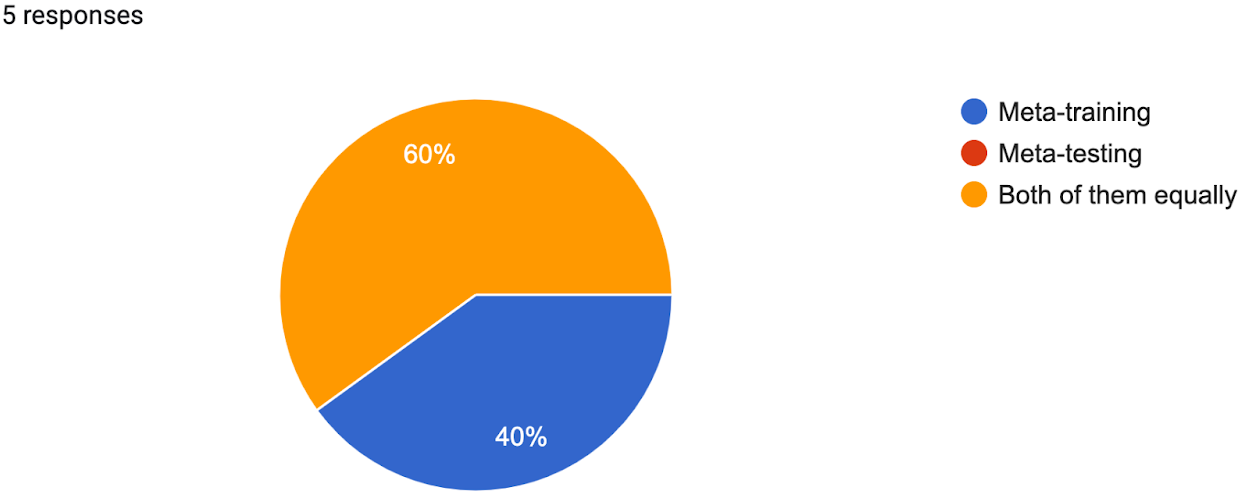}
    \caption{Participants give a slightly higher importance weight to meta-training than meta-testing.}
    \label{fig:my_label}
\end{figure}

\ul{\textbf{Question 18: Did you build an algorithm ranking?}}
\begin{figure}[H]
    \centering
    \includegraphics[width=0.7\textwidth]{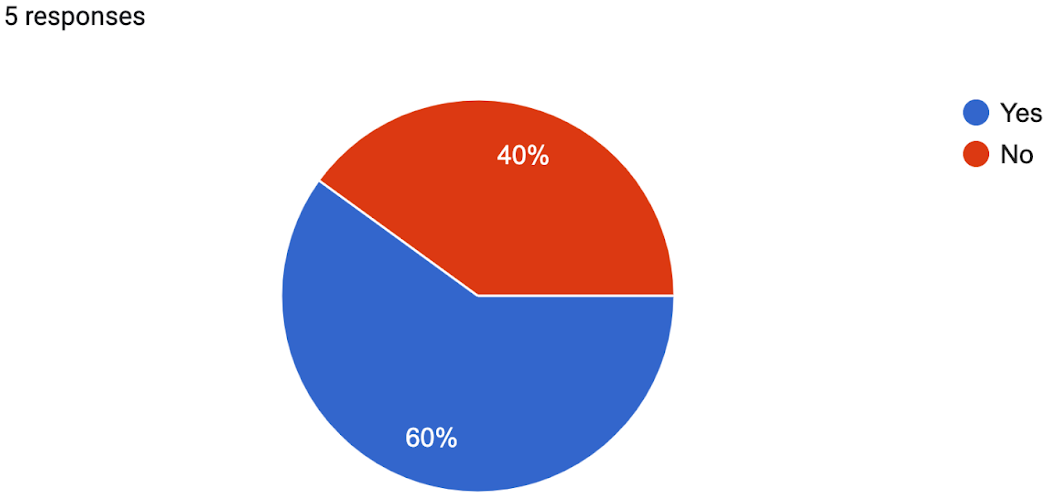}
    \caption{More than half of the participants build an algorithm ranking and use it as a tool for selecting algorithms.}
    \label{fig:my_label}
\end{figure}

\ul{\textbf{Question 19: Did you use Reinforcement Learning to train your agent?}}
\begin{figure}[H]
    \centering
    \includegraphics[width=0.7\textwidth]{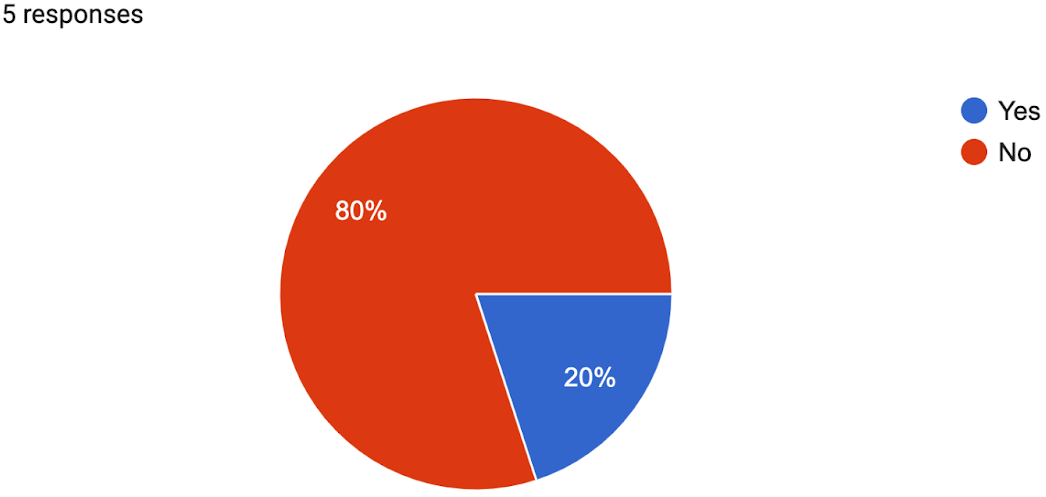}
    \caption{Only one participant applies Reinforcement Learning to train the agent.}
    \label{fig:my_label}
\end{figure}

\subsection{METHOD IMPLEMENTATION}

\ul{\textbf{Question 20: What is the percentage of originality of your method/implementation?}}
\begin{figure}[H]
    \centering
    \includegraphics[width=1\textwidth]{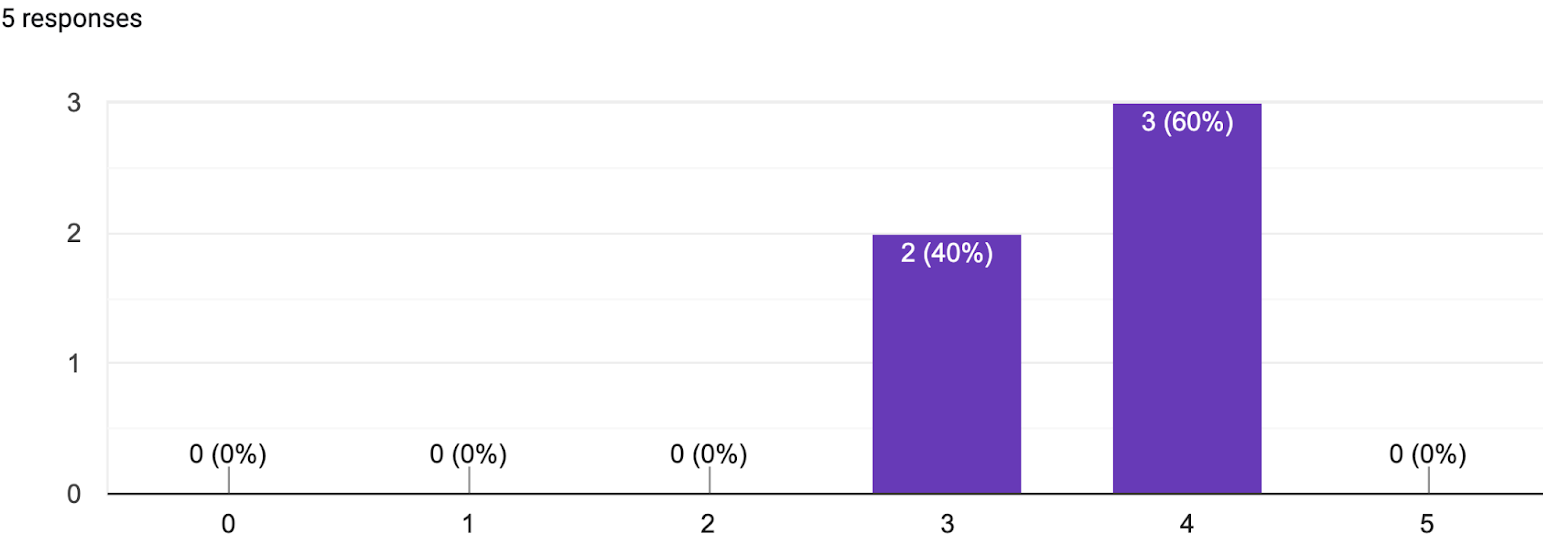}
    \caption{The percentage of originality of their methods/implementations ranges from 60\% to 80\%.}
    \label{fig:my_label}
\end{figure}

\ul{\textbf{Question 21: Did you use a pre-trained agent?}}
\begin{figure}[H]
    \centering
    \includegraphics[width=0.7\textwidth]{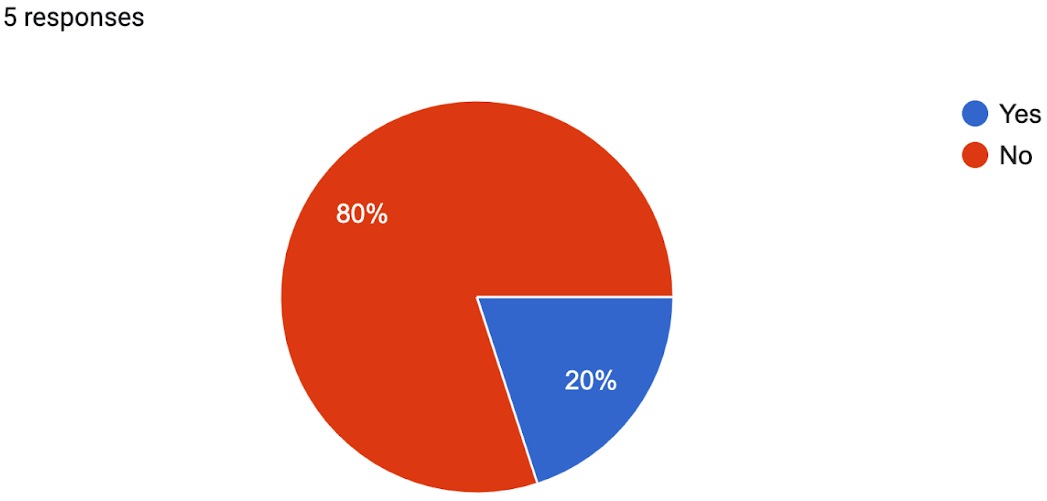}
    \caption{Only one participant pre-train their agent using all the provided datasets in meta-training.}
    \label{fig:my_label}
\end{figure}

\ul{\textbf{Question 22: Is your method strongly based on an existing solution? Which one(s)?}}
\begin{figure}[H]
    \centering
    \includegraphics[width=1\textwidth]{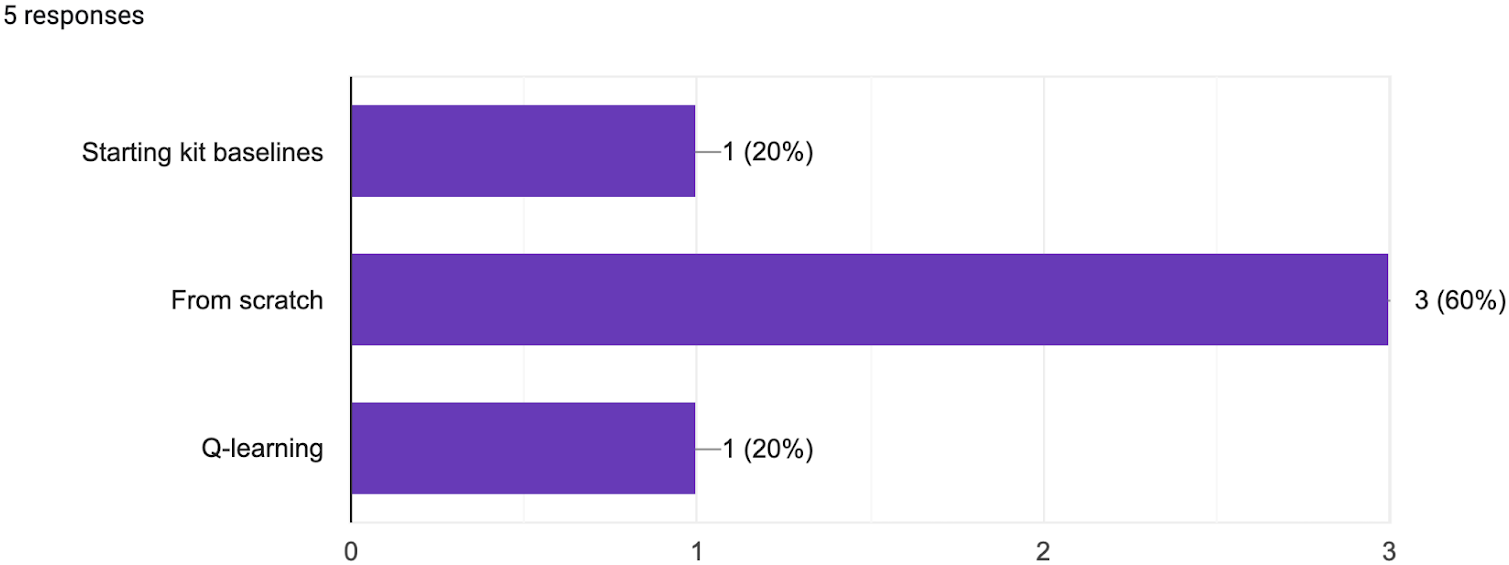}
    \caption{Q-learning and starting kit baselines served as bases for 2 methods, while the other methods were built from scratch.}
    \label{fig:my_label}
\end{figure}

\ul{\textbf{Question 23: Did you use Neural Networks for your agent?}}
\begin{figure}[H]
    \centering
    \includegraphics[width=0.7\textwidth]{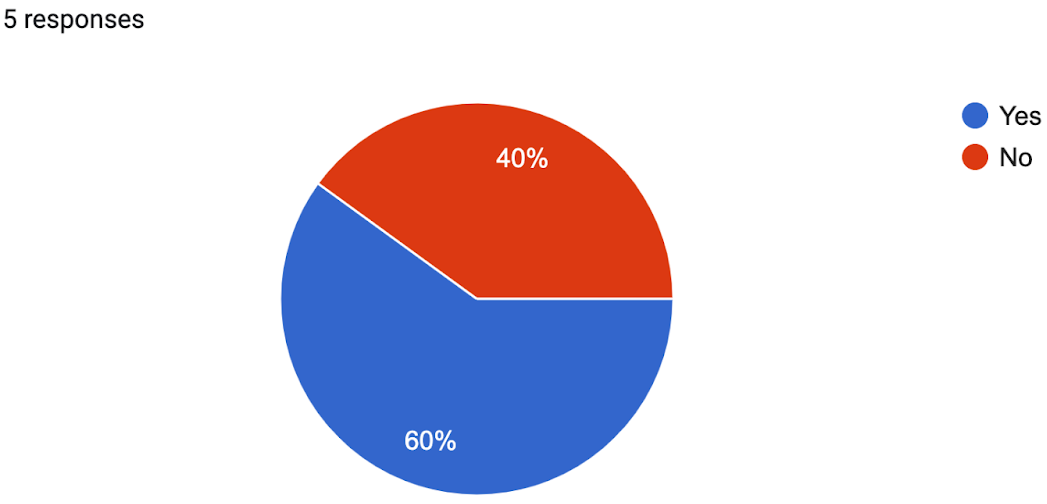}
    \caption{Neural Networks were implemented in 3 out of 5 methods.}
    \label{fig:my_label}
\end{figure}

\ul{\textbf{Question 24: Do you find the provided libraries / packages / frameworks sufficient?}}
\begin{figure}[H]
    \centering
    \includegraphics[width=0.7\textwidth]{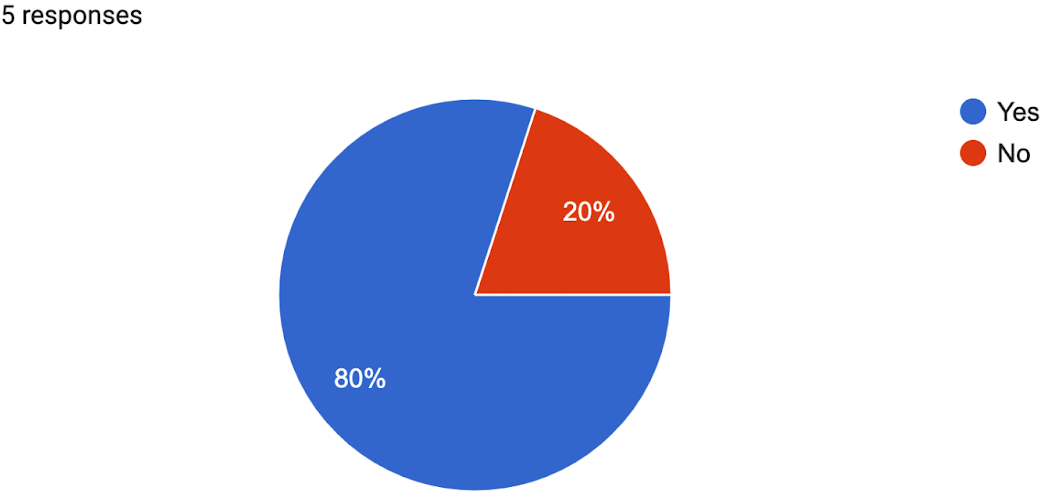}
    \caption{The provided libraries/packages/frameworks were sufficient for most of the participants.}
    \label{fig:my_label}
\end{figure}

\ul{\textbf{Question 25: Check all Python packages/frameworks you used.}}
\begin{figure}[H]
    \centering
    \includegraphics[width=1\textwidth]{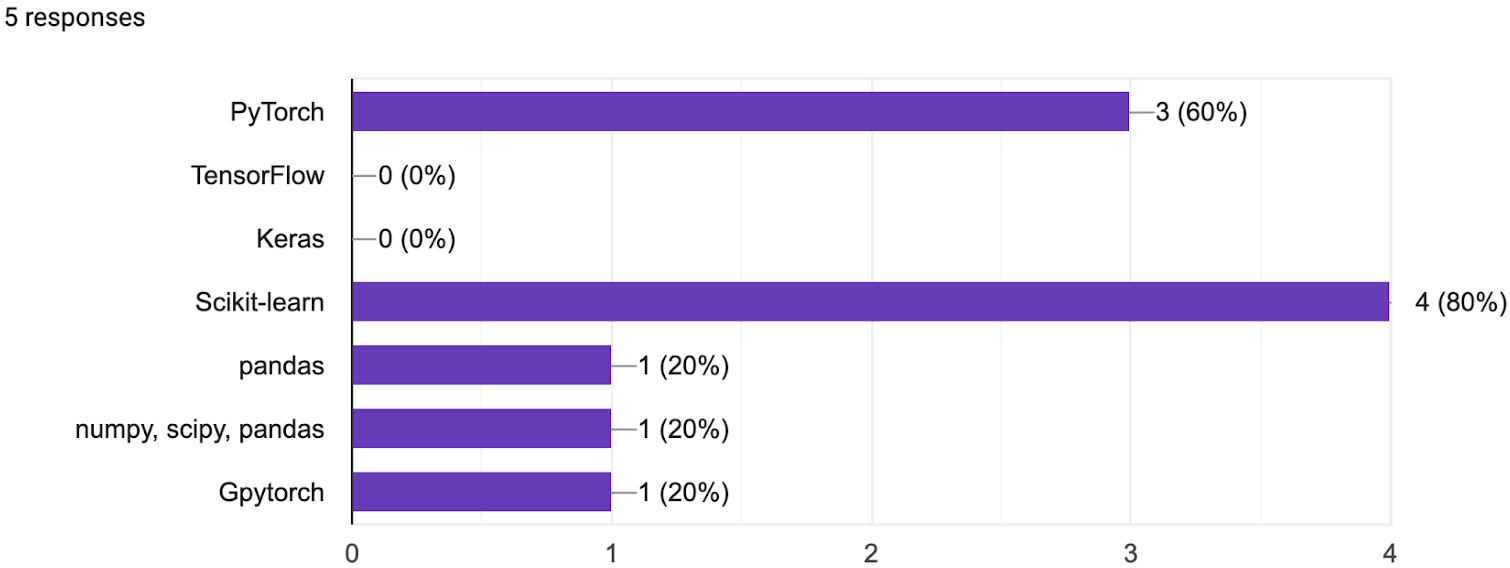}
    \caption{Scikit-learn and Pytorch are the most used packages by participants.}
    \label{fig:my_label}
\end{figure}

\ul{\textbf{Question 26: Did you use any specific AutoML / Meta-learning / Hyperparameter Optimization libraries?}}
\begin{figure}[H]
    \centering
    \includegraphics[width=0.7\textwidth]{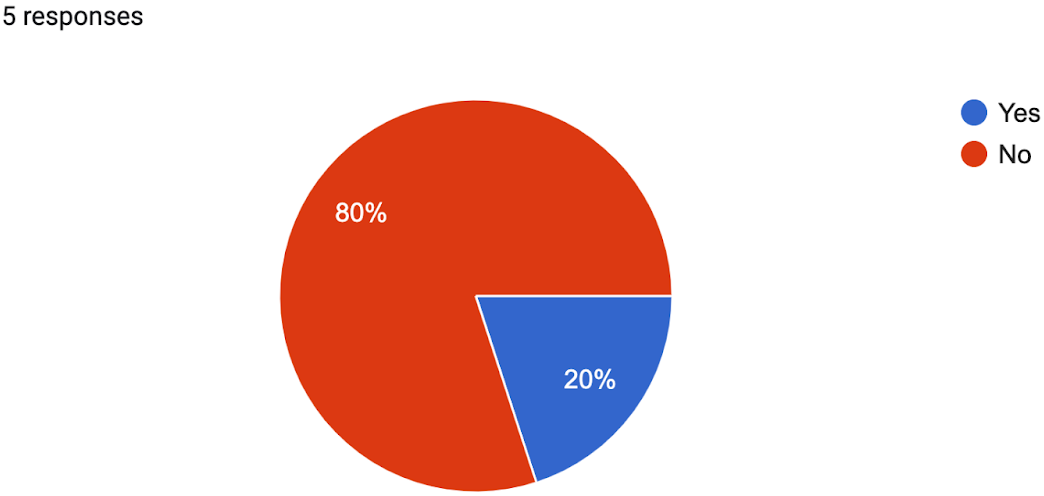}
    \caption{Only one participant uses SMAC for  hyperparameter optimization.}
    \label{fig:my_label}
\end{figure}

\ul{\textbf{Question 27: Was it difficult for you to deal with the provided data format of the learning curves and meta-features?}}
\begin{figure}[H]
    \centering
    \includegraphics[width=0.8\textwidth]{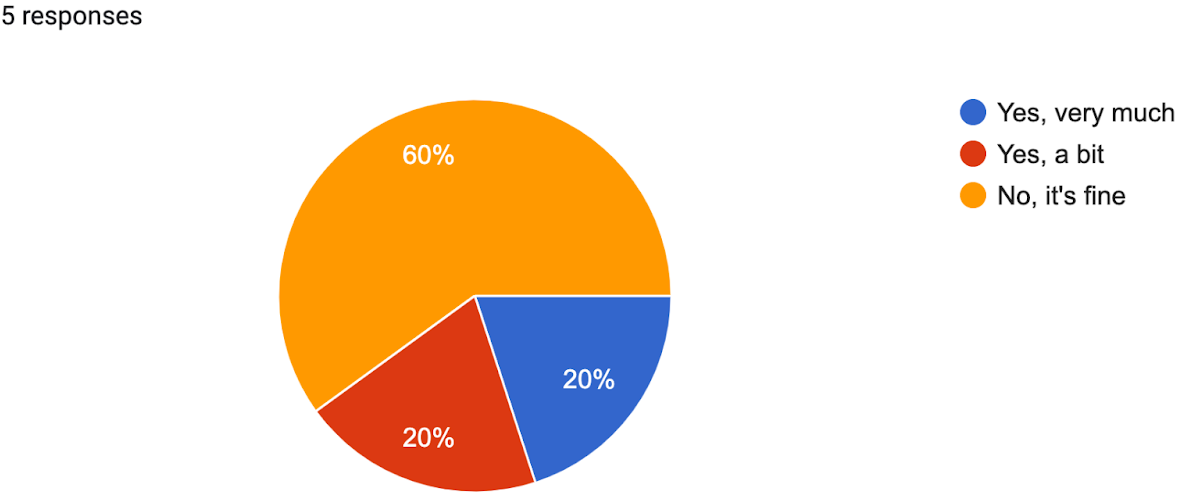}
    \caption{40\% of participants struggled with the provided data format. Some comments include: rather than nested dictionaries create a single dictionary with a tuple identifier (dataset, algorithm).}
    \label{fig:my_label}
\end{figure}

\ul{\textbf{Question 28: How much time did you spend developing your agents?}}
\begin{figure}[H]
    \centering
    \includegraphics[width=0.9\textwidth]{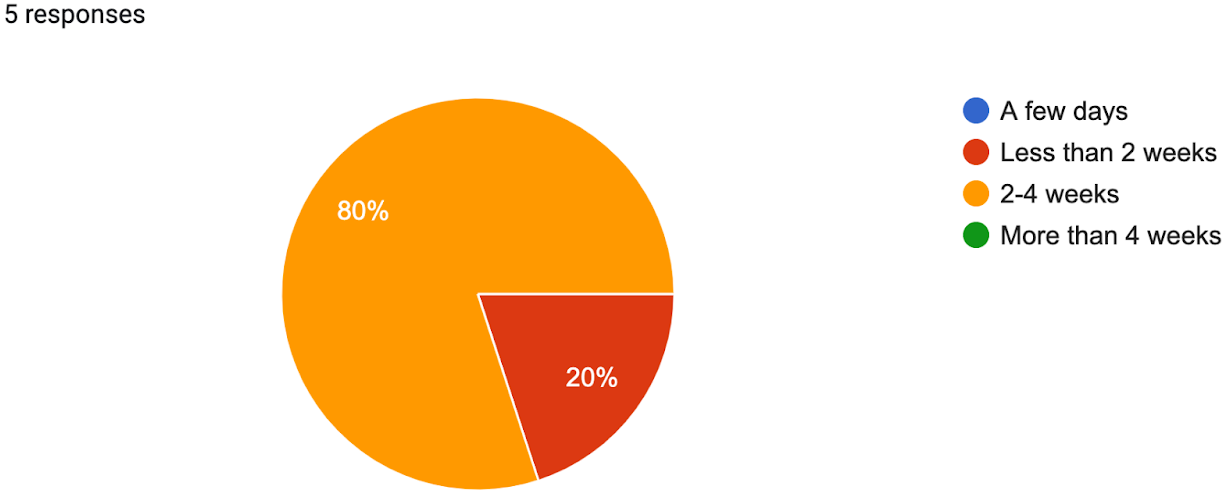}
    \caption{All participants completed their solutions within 4 weeks.}
    \label{fig:my_label}
\end{figure}

\ul{\textbf{Question 29: What's the difficulty induced by the computation resource (memory, time budget, etc) constraints?}}
\begin{figure}[H]
    \centering
    \includegraphics[width=0.8\textwidth]{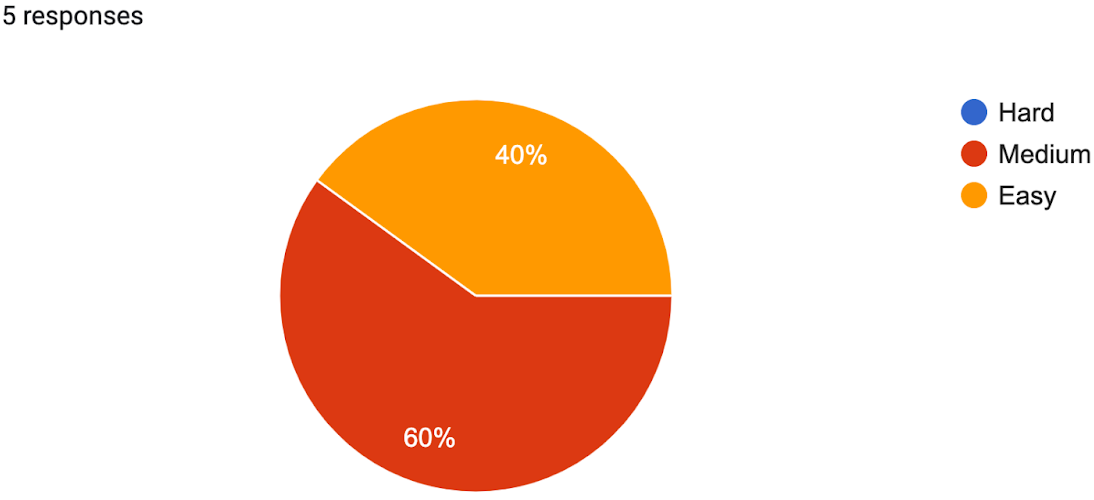}
    \caption{The provided computation resource was reasonable to participants.}
    \label{fig:my_label}
\end{figure}

\subsection{USER EXPERIENCE}

\ul{\textbf{Question 30: Was the challenge duration enough for you to develop your methods?}}
\begin{figure}[H]
    \centering
    \includegraphics[width=0.9\textwidth]{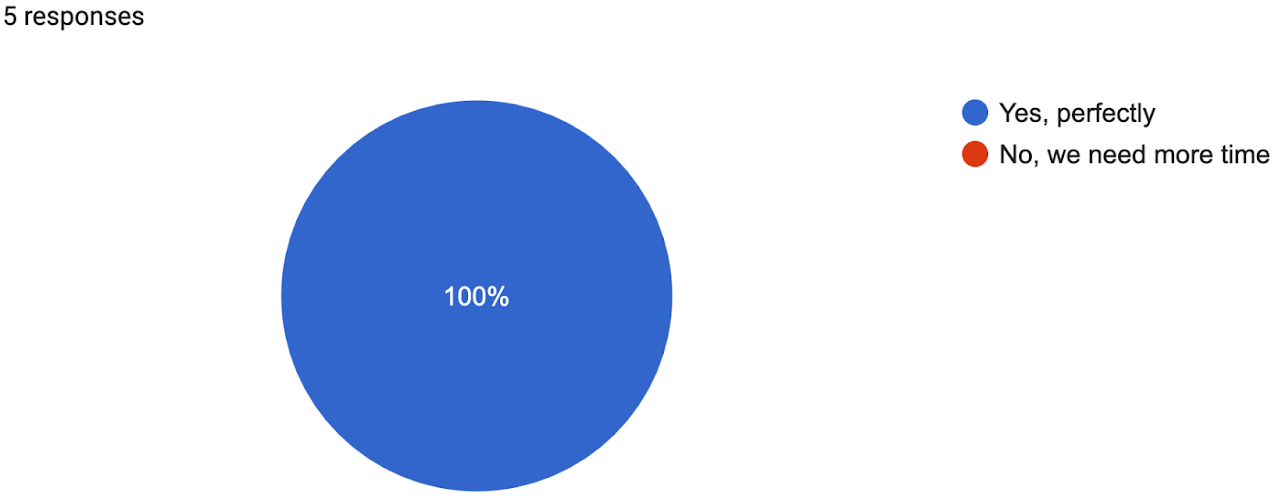}
    \caption{The challenge duration was enough for the participants.}
    \label{fig:my_label}
\end{figure}

\ul{\textbf{Question 31: Your evaluation on the starting kit}}
\begin{figure}[H]
    \centering
    \includegraphics[width=1\textwidth]{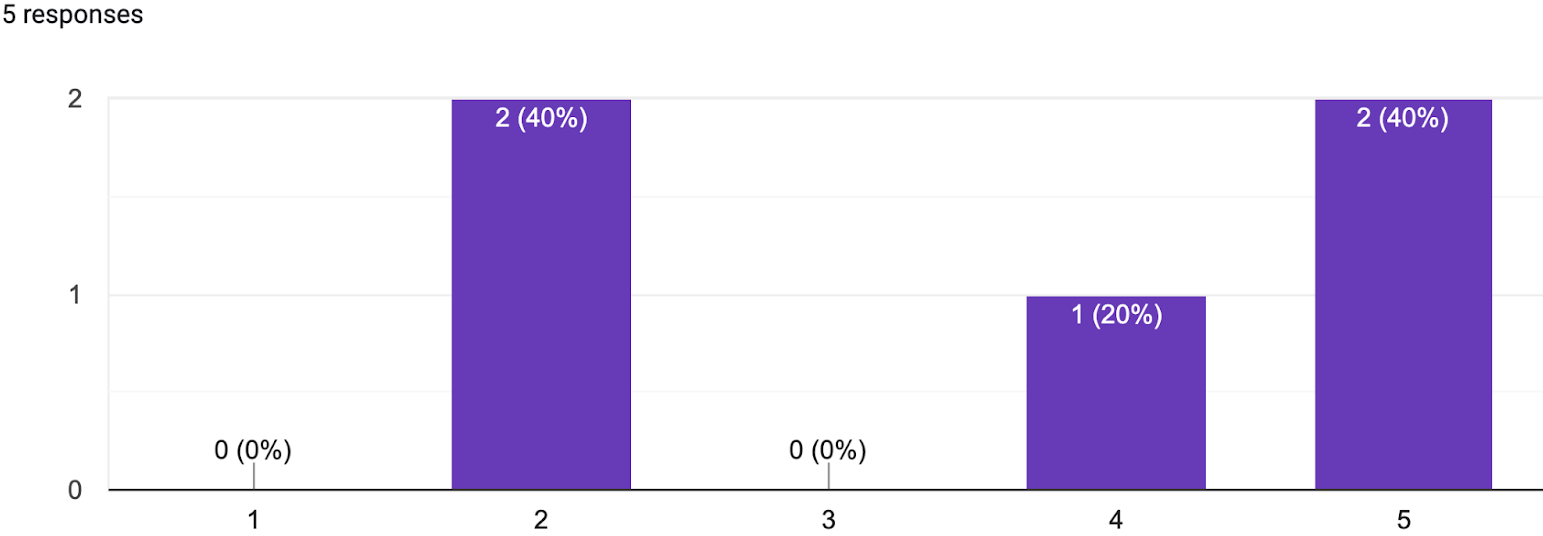}
    \caption{Some improvements on the starting kit need to be made (see below).}
    \label{fig:my_label}
\end{figure}

\ul{\textbf{Question 32: Which improvements can be made for the starting kit? You are welcome to list all issues and bugs you ran into.}}
\begin{enumerate}
    \item The algorithm meta features were non-informative, the dataset meta features' categorical columns at some point were non informative and the categories were not entirely represented in the validation dataset. Also the environment allows to exploit 0 budgets to improve the agent's performance in a non realistic way. Also we wanted to use many libraries, that were not available (e.g pytorch geometric or networkx) or with other versions (e.g. sklearn's quantile regression) was not available initially. The validation \& test set (on server) were slightly different. In particular, the distribution of timestamps was dramatically different.

    \item Return the whole learning curve after a suggestion, not only the last observed budget and the last observed performance.
    
    \item It may be useful to query the learning curve at specific iterations or explain clearly the meaning of the timestamps of the observed samples. If they are random or there is not an underlying pattern or structure, it is difficult to predict the next budget.
    
    \item The final rank should be obtained by running the methods on other meta-train dataset. In the current set-up, the test curve is highly correlated to the validation curve (seen in the development stage), therefore simply overfitting the latter will probably help to obtain good results in the former.
    
    \item Decreasing the budget when the suggested algorithm+budget tuple is not enough to query a new point in the learning curve is unrealistic. In the real world, once we decide to run some algorithm for a given time-range or number of epochs, we will obtain some change in the accuracy (unless the learning curve is in a plateau).
\end{enumerate}

\ul{\textbf{Question 33: Your evaluation on the challenge website:\\ https://codalab.lisn.upsaclay.fr/competitions/753}}
\begin{figure}[H]
    \centering
    \includegraphics[width=1\textwidth]{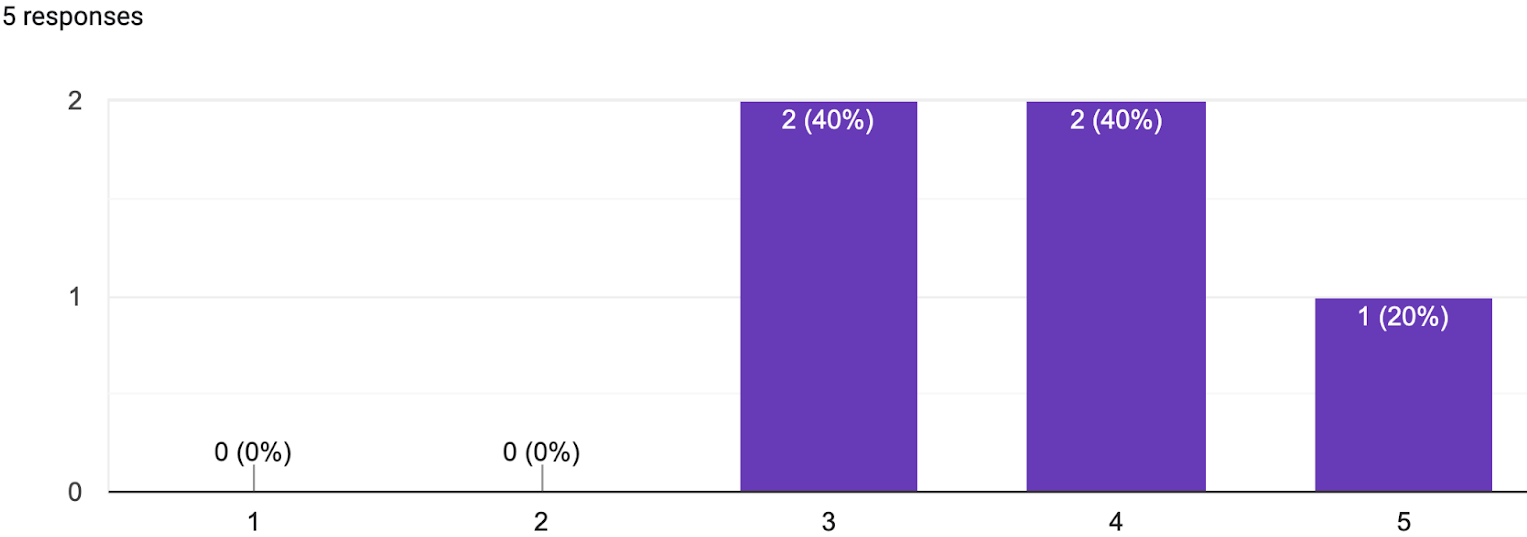}
    \caption{One comment for the challenge website is: the way the ALC is explained is not very straightforward.}
    \label{fig:my_label}
\end{figure}

\end{document}